\documentclass[manuscript,natbib=False]{acmart}
\AtBeginDocument{%
  }

\copyrightyear{2026}
\acmYear{2026}
\setcopyright{cc}
\setcctype{by-nc-nd}
\acmConference[SAC '26]{The 41st ACM/SIGAPP Symposium on Applied Computing}{March 23--27, 2026}{Thessaloniki, Greece}
\acmBooktitle{The 41st ACM/SIGAPP Symposium on Applied Computing (SAC '26), March 23--27, 2026, Thessaloniki, Greece}
\acmPrice{}

\RequirePackage[
  datamodel=acmdatamodel,
  style=acmnumeric,
  ]{biblatex}

\usepackage{hyperref}
\usepackage{url}
\usepackage{booktabs}       
\usepackage{amsfonts}       
\usepackage{nicefrac}       
\usepackage{microtype}      
\usepackage{xcolor}         
\usepackage{amsmath}
\usepackage{qcircuit} 
\usepackage{graphicx}
\usepackage{adjustbox}
\usepackage{caption}
\usepackage{subcaption}
\usepackage{algorithm}
\usepackage{algpseudocode}
\usepackage{colortbl}
\usepackage{placeins}
\usepackage{tikz}
\usetikzlibrary{tikzmark}
\usepackage{float} 

\begin{document}

\title{SEval-NAS: A Search-Agnostic Evaluation for Neural Architecture Search}
\renewcommand{\shorttitle}{SEval-NAS: A Search-Agnostic Evaluation for Neural Architecture Search}



 \author{Atah Nuh Mih}
 \affiliation{%
   \institution{Analytics Everywhere Lab, \\University of New Brunswick}
   \city{Fredericton}
   \country{Canada}}
 \email{atah.mih@unb.ca}
 \orcid{0009-0001-4931-7207}

 \author{Jianzhou Wang} 
 \affiliation{%
   \institution{Analytics Everywhere Lab, \\University of New Brunswick}
   \city{Fredericton}
   \country{Canada}}
 \email{maxwell.wang@unb.ca}
 \orcid{0009-0000-2833-1598}
 
 \author{Truong Thanh Hung Nguyen}
 \affiliation{%
   \institution{Analytics Everywhere Lab, \\University of New Brunswick}
   \city{Fredericton}
   \country{Canada}}
 \email{hung.ntt@unb.ca}
 \orcid{0000-0002-6750-9536}

 \author{Hung Cao}
 \affiliation{%
   \institution{Analytics Everywhere Lab, \\University of New Brunswick}
   \city{Fredericton}
   \country{Canada}}
 \email{hcao3@unb.ca}
 \orcid{0000-0002-0788-4377}








\begin{abstract}
Neural architecture search (NAS) automates the discovery of neural networks that meet specified criteria, yet its evaluation procedures are often hardcoded, limiting the ability to introduce new metrics. This issue is especially pronounced in hardware-aware NAS, where objectives depend on target devices such as edge hardware. To address this limitation, we propose SEval-NAS, a metric-evaluation mechanism that converts architectures to strings, embeds them as vectors, and predicts performance metrics. Using NATS-Bench and HW-NAS-Bench, we evaluated accuracy, latency, and memory. Kendall's $\tau$ correlations showed stronger latency and memory predictions than accuracy, indicating the suitability of SEval-NAS as a hardware cost predictor. We further integrated SEval-NAS into FreeREA to evaluate metrics not originally included. The method successfully ranked FreeREA-generated architectures, maintained search time, and required minimal algorithmic changes. Our implementation is available at: \url{https://github.com/Analytics-Everywhere-Lab/neural-architecture-search}.

\end{abstract}



\maketitle

\section{Introduction} \label{sec:introduction}

Neural architecture search (NAS) was developed to automate the design of neural networks (NNs), addressing the knowledge gap required to design these networks manually. 
Traditional NAS \cite{Zoph2016} used reinforcement learning (RL) to generate variable-length strings representing architectures, achieving state-of-the-art performance but at high computational cost.
This shortcoming prompted research into more efficient search methods, such as a cell-based search space with learnable transferable architectures \cite{Zoph2017}, Efficient NAS (ENAS) \cite{Pham2018}, Progressive NAS (PNAS) \cite{LiuC2018}, and Regularized Evolution for Image Classifier Architecture Search \cite{Real2019}.
In general, NAS is divided into five major components: (1) the search space containing predefined operation sets; (2) the controller that determines how neural architectures are generated; (3) the candidate architecture(s) generated; (4) the evaluation phase with a strategy for assessing architectures feasibility; and (5) the optimal architecture(s) that satisfy search objectives.

The evaluation phase is a critical step in NAS because it assesses candidate architectures for the desired objectives and directs the search toward optimal designs \cite{Chitty2023}. Depending on the evaluation method, this phase can impose substantial search cost, as candidate architectures must be trained and tested to estimate performance \cite{Cavagnero2023}. For example, \cite{Zoph2017} trained each candidate to convergence on proxy data, resulting in a search cost of 22,400 GPU hours. The high cost of full training motivated the use of incomplete training to accelerate candidate ranking and reduce search costs \cite{Ren2021}.
Another challenge in existing NAS approaches is their limited flexibility in accommodating new evaluation metrics. Different NAS methods target different performance criteria, and hardware-aware NAS often incorporates metrics such as latency \cite{Sinha2024, WuB2019, Imed2023} and memory \cite{LiY2023} to select architectures suited for platforms like edge devices. As a result, evaluation criteria are typically hardcoded into the search, making it difficult to introduce new metrics without redesigning the algorithm.
Therefore, an evaluation mechanism that is independent of the search algorithm and can be flexibly adapted to any NAS approach is necessary.

To address these challenges, we propose a search-agnostic evaluation approach called SEval-NAS. It converts any NN into its string representation, encodes the string to obtain the vector embedding, and predicts the evaluation metrics. This is based on the premise that NNs' performance reflects the structural dependencies of their internal operations (e.g., type of convolution, number of filters, type of activations). Extracting this structural information can, therefore, help predict their given performance.
We show that SEval-NAS supports different types of metrics and evaluation objectives and can be directly applied to an existing NAS method with minimal changes to the search algorithm and without significantly affecting the search. 
We experiment on two NAS benchmarks: NATS-Bench \cite{Dong2022} and HW-NAS-Bench \cite{LiC2021} for accuracy, latency, and memory, and further assess how our method affects a NAS algorithm (i.e., FreeREA \cite{Cavagnero2023}). 

In summary, this work presents the following contributions:
\begin{itemize}
    \item A network-to-string conversion mechanism that traverses the autograd graph of any NN and generates its textual representation, making it adaptive to all types of NNs. 
    \item An encoder-predictor network (i.e., an evaluator) that extracts meaningful relationships between the strings and their evaluation metric. This network can be designed to include any evaluation metric (notably hardware costs) and any number of evaluation objectives. 
    \item SEval-NAS that is independent of the NAS algorithm and combines the network-to-string conversion mechanism and the evaluator to evaluate candidate architectures in NAS.
    \item An ablation study of three different encoder/decoder models (T5-small, T5-base, and T5-large) in SEval-NAS on NAS benchmarks.
\end{itemize}









\section{Literature Review} \label{sec:lit_rev}




\subsubsection*{\textbf{Training-Free NAS}}
The costly evaluation of candidate architectures has motivated the development of training-free metrics that evaluate candidate architectures and reduce the search time of NAS. \cite{Baker2018} proposed regression models to predict the final performance of models from learning curve trajectories based on features obtained from the neural architectures, hyperparameters, and time-series measurements. 
\cite{Mellor2021} proposed NASWOT, which examines the correlation of activations between data points in untrained NNs and scores networks based on the binary codes corresponding to this correlation. \cite{ChenW2021} proposed TE-NAS, a training-free NAS that analyzes the spectrum of the neural tangent kernel (NTK) and the number of linear regions to rank candidate architectures. While these methods successfully address the evaluation of candidate architectures, they solely focus on accuracy as their performance metrics. We extend beyond accuracy alone by including hardware metrics to assess the suitability of architectures for different computing environments. 

Hardware cost predictors have been developed, such as a 
nn-Meter \cite{ZhangL2021}, a latency predictor for edge devices; and a GPU estimator for deep learning models \cite{GaoY2020}. Integrating different single-purpose cost predictors will increase the complexity of the design, so we propose a multi-purpose cost estimator that can incorporate different hardware costs. 

\subsubsection*{\textbf{Hardware-Aware NAS}}
The search for good neural architectures extends beyond just high-accuracy networks. In cases where the hardware environment is crucial, cost metrics must be considered when evaluating the NNs. This requirement gave rise to HW-NAS, which includes a hardware cost metric and test accuracy in evaluating candidate architectures. Several HW-NAS approaches have been proposed, such as SqueezeNext \cite{Gholami2018}, IRLAS \cite{GuoM2019}, and FB-Net \cite{WuB2019}. These works only provide the optimal architectures obtained from the search, leaving a question about their performance if the search were to be evaluated differently. 
Hardware-aware NAS equally targets edge devices because their hardware environments require NNs suitable for their resource constraints. Several NAS methods have been developed for edge devices \cite{Kundu2023, Sinha2024, LyuB2022, Risso2022}. While these works have been successful in searching optimized NNs, they are usually designed to satisfy a single hardware cost metric. Latency has often been used as the evaluation metric in hardware-aware NAS \cite{LuoX2020, ZhangY2024, Imed2023, LiY2023}, whereas few works include multiple cost metrics in their design \cite{Richey2024}. 
These NAS methods are generally multi-objective and aim to satisfy more than one primary objective pre-defined while designing the search. Our approach provides an easy integration of a cost evaluator for any desired objective (e.g., accuracy, latency, and memory) and several objectives (e.g., single or bi-objectives).


\subsubsection*{\textbf{NAS as a String Search Problem}}
Treating NAS as a string search problem was first explored in \cite{Zoph2017}, where architectures were encoded as variable-length strings generated by a recurrent controller using reinforcement learning. GeNet \cite{Xie2017} followed a similar idea, representing architectures as fixed-length binary strings and applying genetic algorithms to evolve them. Neural Architecture Optimization (NAO) \cite{LuoR2021} introduced an encoder–predictor–decoder framework that embeds architectures into a continuous space, predicts accuracy, and reconstructs architectures, with optimization performed directly in the embedding space. As in many NAS methods, prediction is tightly integrated into the search. In contrast, our prediction mechanism is plug-and-play and can be added to existing NAS pipelines, and we extend prediction to hardware costs, which is essential for HW-NAS. More recent work such as EVOPROMPTING \cite{ChenA2023} uses language modeling and prompting to generate code-level architectures, though it operates at a much higher level by producing program code rather than structured architectural representations.

\section{Methodology} \label{sec:method}

\begin{figure}[]
    \centering
    \includegraphics[width=\linewidth]{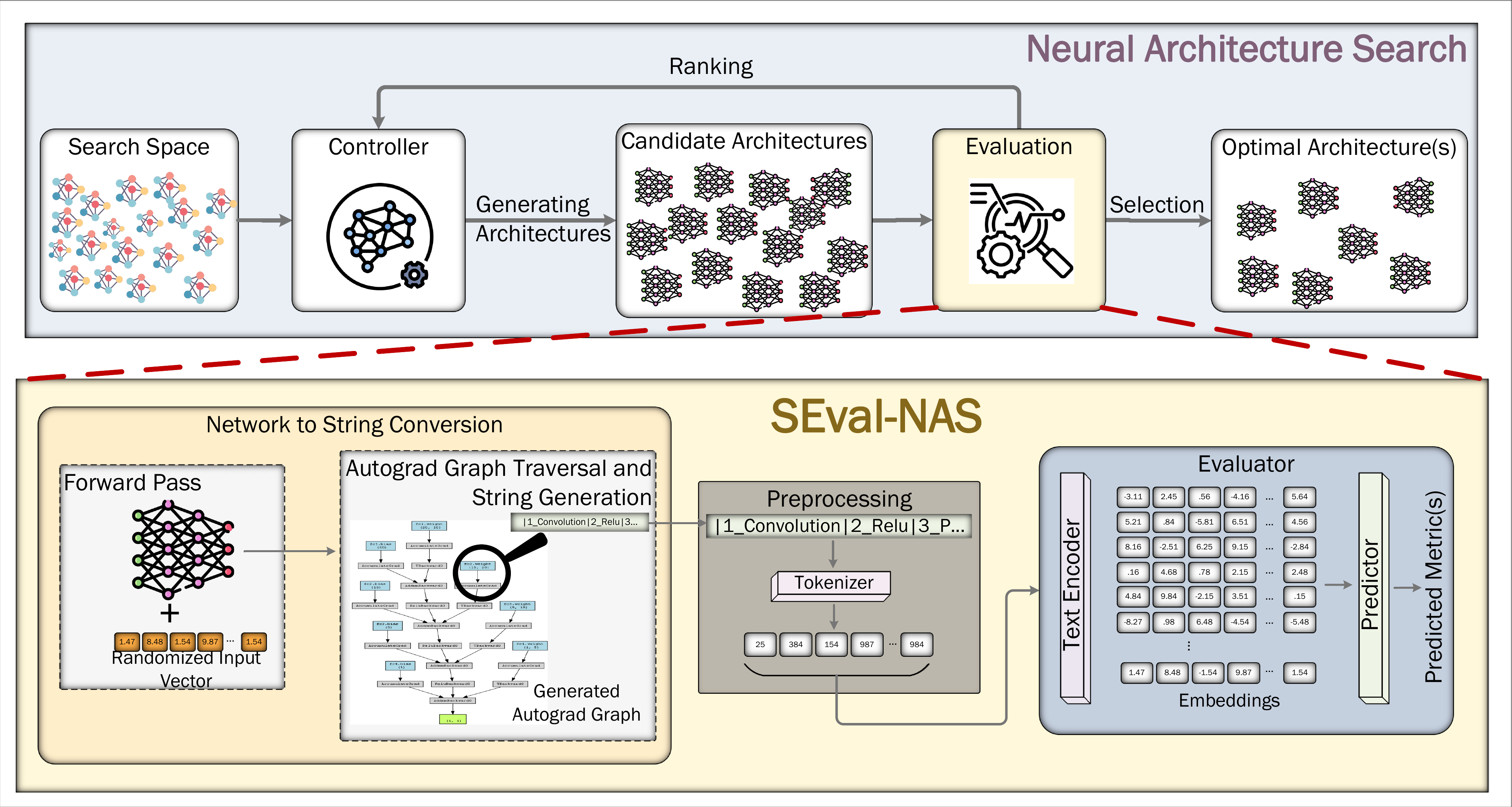}
    \caption{Proposed SEval-NAS methodology and its integration in a NAS pipeline}
    \label{fig:methodology}
\end{figure}

Our proposed SEval-NAS framework is designed to evaluate neural architectures within a NAS pipeline by leveraging a formalized string-based representation and a predictive evaluation model. Let $\mathcal{A}$ denote the set of candidate neural architectures, where each architecture $a \in \mathcal{A}$ is characterized by its computational graph. The methodology transforms each architecture into a standardized string representation, which is subsequently processed by an evaluator to predict performance metrics. The predicted metrics guide the NAS controller in optimizing the search process. The framework consists of two primary components: (1) a network-to-string conversion mechanism and (2) an evaluator network for performance prediction.
Fig. \ref{fig:methodology} provides a schematic of the proposed methodology, illustrating its integration within a NAS pipeline.

\subsection{Net-to-String Conversion}


The network-to-string conversion process maps a neural architecture $a \in \mathcal{A}$ to a string representation $s_a \in \mathcal{S}$, where $\mathcal{S}$ is the space of all possible string representations. Let $G_a = (V_a, E_a)$ represent the computational graph of architecture $a$ (generated during the forward pass), with vertices $V_a$ corresponding to operations (e.g., convolution, pooling, ReLU) and edges $E_a$ representing data flow between operations. The conversion function $f: \mathcal{A} \to \mathcal{S}$ traverses $G_a$ (breadth-first) to extract structural and operational details, yielding a string $s_a$ that encapsulates the architecture's configuration.

Formally, the conversion process is defined as:
\(
s_a = f(G_a)
\), where $f$ systematically traverses $V_a$ and $E_a$ to encode operations and their connectivity into a standardized format. The resulting string $s_a$ is tokenized into a sequence of tokens $T_a = \{t_1, t_2, \dots, t_n\}$, where each token $t_i$ corresponds to a specific operation or parameter in the computational graph. This tokenization ensures a universal and consistent representation, enabling compatibility across diverse NAS tasks and datasets. 
The conversion is described in Alg.~\ref{alg:net_conversion}.
\begin{algorithm}[]
\caption{Network to String Conversion}\label{alg:net_conversion}
\small
\textbf{Input:} \text{Neural Network: $net_a$; Input Tensor: $inp$} 
\hspace*{\fill} \\
\textbf{Output:} \text{Output token $T_a$}
\hspace*{\fill}
\begin{algorithmic}[1]
    \State $G_a \longleftarrow net_a(inp)$ \Comment{forward pass}
    \State $node = G_a(root)$ \Comment{root node}
    \State $s_a = \varnothing$ \Comment{output string}
    \State $seen = \varnothing$ \Comment{keep track of traversal}
    \State $node\_id = 0$ \Comment{node index}
    \Function{get\_string}{node}\
        \If {$node$ in $seen$}
            \Return
        \EndIf
        \State $seen \longleftarrow node$
        \State $next\_nodes = node.next\_functions$ \Comment{neighbouring vertices}
        \For {$u$ in $next\_nodes$}
            \State \Call{get\_string}{u}
        \EndFor
        
        \State $vars = node.variable$ \Comment{get operations and variables}
        \State $name = node.name$
        \State $s_a \longleftarrow name, node\_id, vars$
    \EndFunction
    \State $T_a = tokenize(s_a)$ \Comment{tokenization with any desired method}
\end{algorithmic}
\end{algorithm}



\subsection{Evaluator}
The tokenized input is processed in the evaluator module to predict its performance metrics. The evaluator module, denoted as $\mathcal{E}$, predicts performance metrics for a given architecture based on its tokenized representation $T_a$. Let $\mathcal{M} = \{m_1, m_2, \dots, m_k\}$ represent the set of target performance metrics (e.g., accuracy, latency, memory usage). The evaluator maps the tokenized input to a vector of predicted metrics:
\(
\hat{m}_a = \mathcal{E}(T_a)
\), where $\hat{m}_a = [\hat{m}_{a,1}, \hat{m}_{a,2}, \dots, \hat{m}_{a,k}] \in \mathbb{R}^k$ denotes the predicted values for the $k$ metrics. 

The module consists of two components: an encoder and a predictor. 
\begin{enumerate}
    \item \textbf{Encoder}: The encoder extracts a high-dimensional vector representation of the architecture, capturing its structural and contextual information. It is represented by a function $g: \mathcal{T} \to \mathbb{R}^d$ that transforms the tokenized sequence $T_a$ into a high-dimensional embedding $e_a \in \mathbb{R}^d$, capturing structural and contextual features of the architecture:
    \[
    e_a = g(T_a).
    \]
    The encoder employs a transformer-based architecture to model dependencies within the token sequence, ensuring robust feature extraction.

    \item \textbf{Predictor}: A function $h: \mathbb{R}^d \to \mathbb{R}^k$ that maps the embedding $e_a$ to the predicted metrics:
    \[
    \hat{m}_a = h(e_a).
    \]
    The prediction layer is a fully connected neural network with $k$ output neurons, where the number of neurons corresponds to the number of target metrics. For single-objective prediction (e.g., latency), $k=1$, while for multi-objective prediction (e.g., accuracy and latency), $k \geq 2$.
\end{enumerate}

The evaluator is trained on a dataset $\mathcal{D} = \{(a_i, m_i)\}_{i=1}^N$, where $m_i \in \mathbb{R}^k$ are the true performance metrics for architecture $a_i$. The training objective minimizes the loss function:
\[
\mathcal{L} = \frac{1}{N} \sum_{i=1}^N \ell(\mathcal{E}(T_{a_i}), m_i),
\]
where $\ell$ is a regression loss (e.g., mean squared error) that measures the discrepancy between predicted and true metrics.

\subsection{Integration into NAS Pipeline}
The SEval-NAS framework is integrated into a NAS pipeline by evaluating candidate architectures generated by the controller. Let $\mathcal{C}$ denote the controller, which generates architectures $a \in \mathcal{A}$ based on a search strategy. The evaluator provides feedback in the form of predicted metrics $\hat{m}_a$, enabling the controller to optimize the search objective:
\[
a^* = \arg\max_{a \in \mathcal{A}} u(\hat{m}_a),
\]
where $u: \mathbb{R}^k \to \mathbb{R}$ is a utility function that aggregates the predicted metrics (e.g., a weighted sum for multi-objective optimization). The schematic of this integration is illustrated in Figure 1, highlighting the closed-loop interaction between the controller, candidate architectures, and the SEval-NAS evaluator.

This modular design ensures that SEval-NAS can be seamlessly incorporated into existing NAS frameworks, such as FreeREA, without requiring significant modifications to the search algorithm. The flexibility of the prediction layer allows adaptation to varying numbers of objectives, enhancing the applicability of SEval-NAS across diverse hardware and performance constraints.









\section{Experiments and Results} \label{sec:experiments}
We evaluated the effectiveness of SEval-NAS using two NAS benchmarks: NATS-Bench \cite{Dong2022} and HW-NAS-Bench \cite{LiC2021}. Our evaluation focused on how well our method's predictions correlated with actual performance metrics. Additionally, we demonstrate the adaptability of our approach by applying it to FreeREA.
We run our experiments on a 13th Gen Intel(R) Core(TM) 9-13900K server equipped with NVIDIA GeForce RTX 4090.


\subsection{Model Configuration}
The evaluator in SEval-NAS is a transformer encoder whose input is the neural network's text representation and outputs the embedding into a regression head for prediction. Specifically, we use the encoder from the T5 transformer \cite{Raffel2020}. It consists of stacked layers containing a self-attention layer and a small feed-forward network, followed by layered normalization and a residual skip connection. Dropout is strategically applied to the feed-forward network, the skip connection, the attention weights, and the stack's input and output. We use three different sizes of T5 models : 
\begin{itemize}
  \item T5-small, which uses 8-headed attention, has only 6 layers each in the encoder and decoder, and has roughly 60 million parameters. 
  \item T5-base, which uses 12-headed attention, has 12 layers each in the encoder and decoder, and has nearly 220 million parameters.
  \item T5-large, which uses 16-headed attention, has 24 layers each in the encoder and decoder, and has approximately 770 million parameters
\end{itemize}
The predictor is a single dense layer whose number of output neurons depends on the number of desired objectives. 


\subsection{Training}
The evaluator (T5-small model) is trained to predict performance metrics on datasets containing NNs and their reported metrics. Other models (T5-base and T5-large) are trained and evaluated via an ablation study (see Appendix \ref{appd:A}). In NAS, these datasets exist as NAS benchmarks containing thousands of neural architectures and metrics, i.e., accuracy, latency, and FLOPS obtained from training and inferencing those networks. Typically, these benchmarks exclude memory usage, prompting the need for additional profiling. To address this, we build a lookup table containing peak memory usage using the built-in PyTorch memory profiling tool (i.e., \textit{torch.cuda.max\_memory\_allocated}).
This tool measures the peak memory allocated to tensors during training, providing an accurate assessment of NNs' memory footprint. The profiling is isolated from memory used by external factors such as libraries or system variables, ensuring precise measurement. 
We train our evaluator on 2 NAS benchmarks: NATS-Bench \cite{Dong2022} and HW-NAS-Bench \cite{LiC2021}. 



\subsection{Experiment 1: Feasibility Testing (Evaluation on NATS-Bench)}
NATS-Bench \cite{Dong2022} is a unified benchmark dataset for searching on both architecture topology and size. It consists of 15,625 different architectures for the Topology Search Space (TSS) and 32,768 architectures for the Size Search Space (SSS) evaluated on CIFAR-10, CIFAR-100, and ImageNet16-120. 

In the TSS, each architecture corresponds to a different cell represented as a densely connected directed acyclic graph (DAG) with four nodes and edges corresponding to operations from a predefined set of 5 operations. For each architecture in the TSS, the cells are stacked 5 times, with output channels set to 16, 32, and 64 for three stages. The search results in a search space containing 15,625 possible architectures configured for the image dataset considered (i.e., CIFAR-10, CIFAR-100 \cite{Krizhevsky2009}, and ImageNet16-120 \cite{Chrabaszcz2017}). 
The SSS searches for architectures by varying the number of channels in each layer (convolution, cell, or block). Each architecture consists of a stacked cell, and the number of channels in each layer is selected from the set $\{8, 16, 24, 32, 40, 48, 56, 64\}$, resulting in 32,768 architectures.
For our experiment, we evaluate the TSS and SSS. In each search space, we separately train the evaluator on the CIFAR-10, CIFAR-100, and ImageNet16-120 architectures, configuring our network for both \textit{(accuracy, memory)} and \textit{(accuracy, latency)} bi-objective setups.





\begin{figure}[htbp]
    \centering
    \begin{minipage}{0.05\linewidth}
        \rotatebox{90}{\fontfamily{phv}\selectfont\large\textbf{\textcolor{violet}{Memory}}} 
    \end{minipage}
    \hspace{0pt}
    \begin{minipage}{0.9\linewidth}
        \begin{subfigure}[b]{0.3\linewidth}
            \centering
            \includegraphics[width=\linewidth]{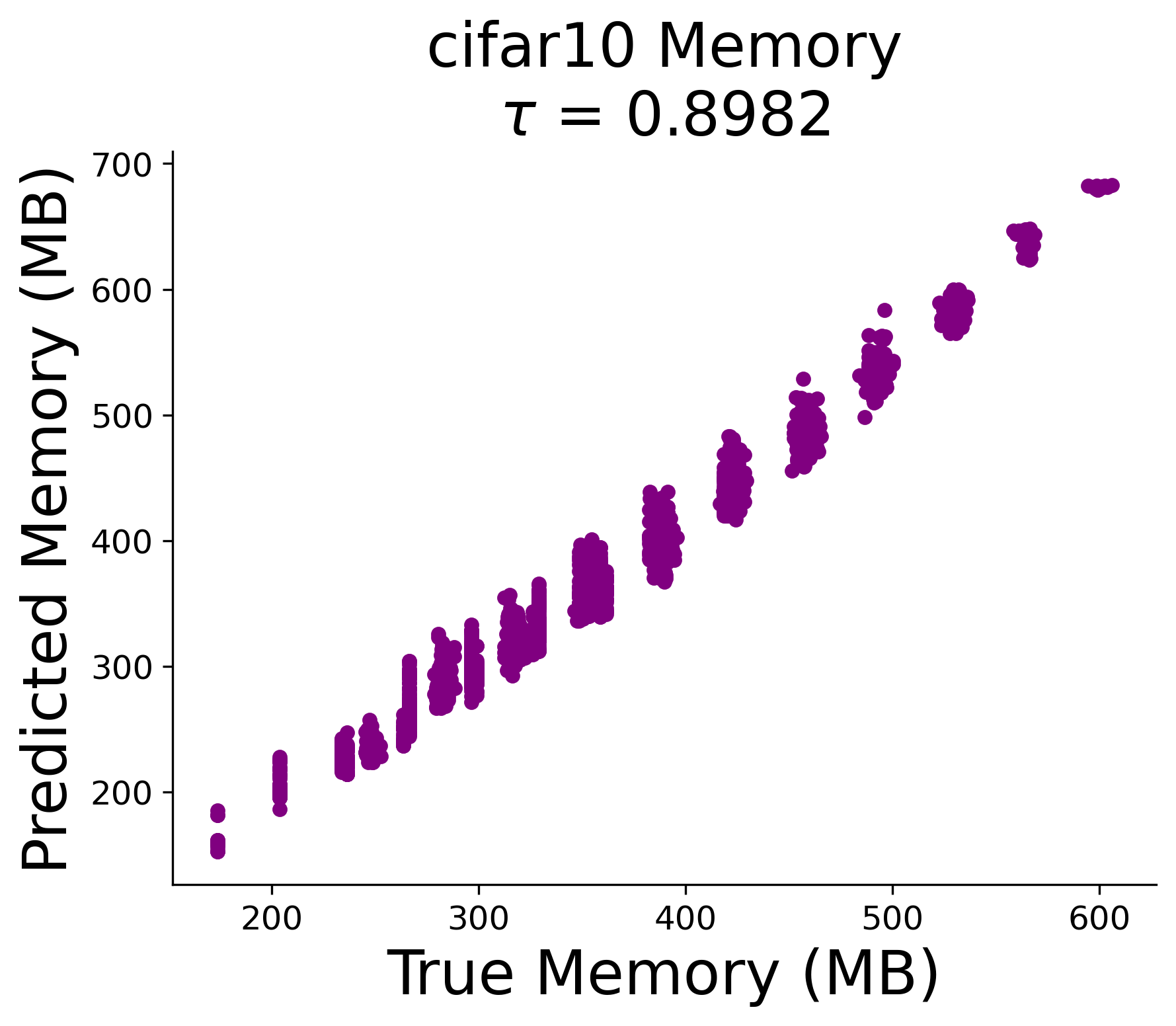}
            \caption{CIFAR-10}
            \label{fig:subfig1}
        \end{subfigure}
        \hfill
        \begin{subfigure}[b]{0.3\linewidth}
            \centering
            \includegraphics[width=\linewidth]{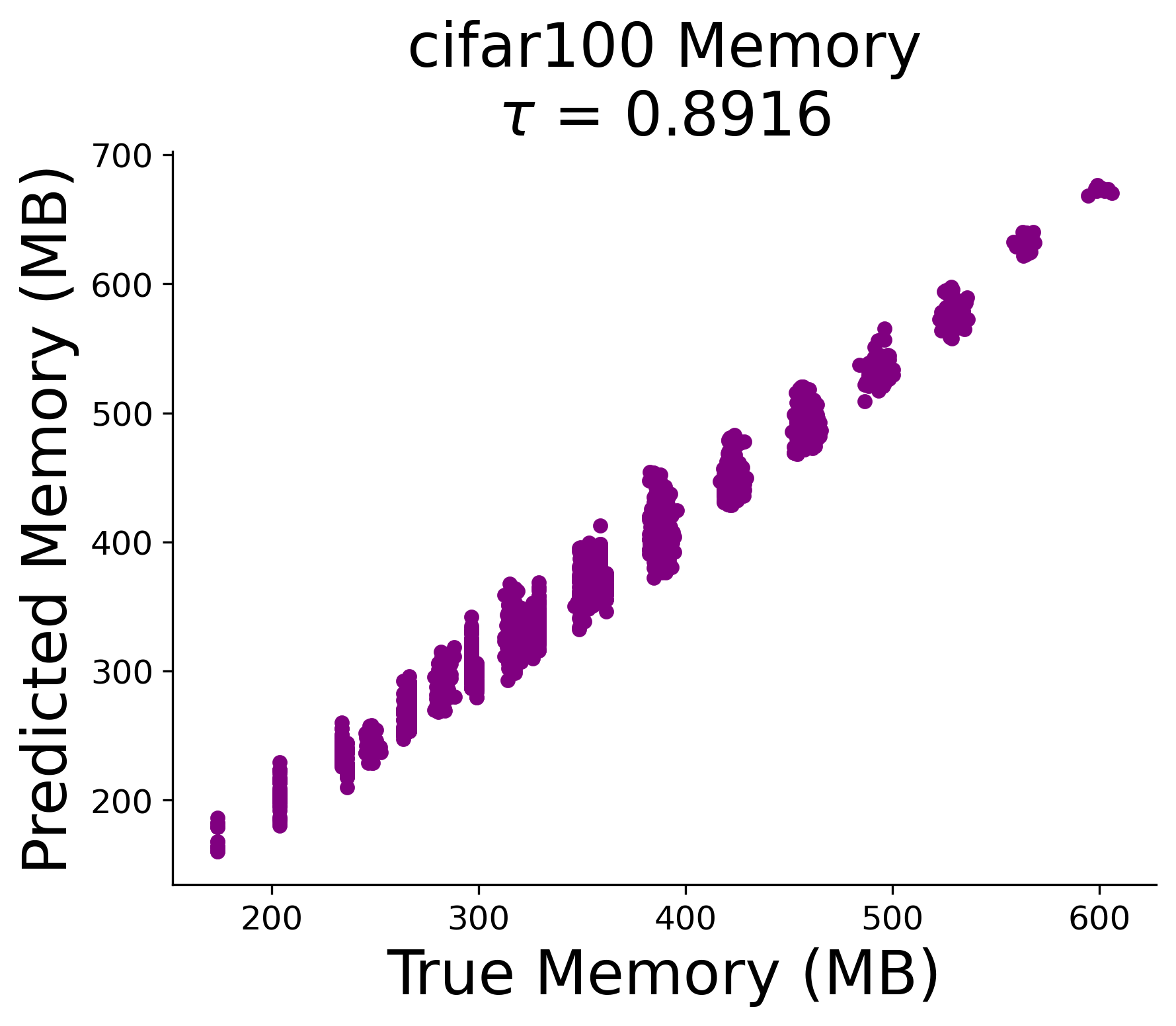}
            \caption{CIFAR-100}
            \label{fig:subfig2}
        \end{subfigure}
        \hfill
        \begin{subfigure}[b]{0.3\linewidth}
            \centering
            \includegraphics[width=\linewidth]{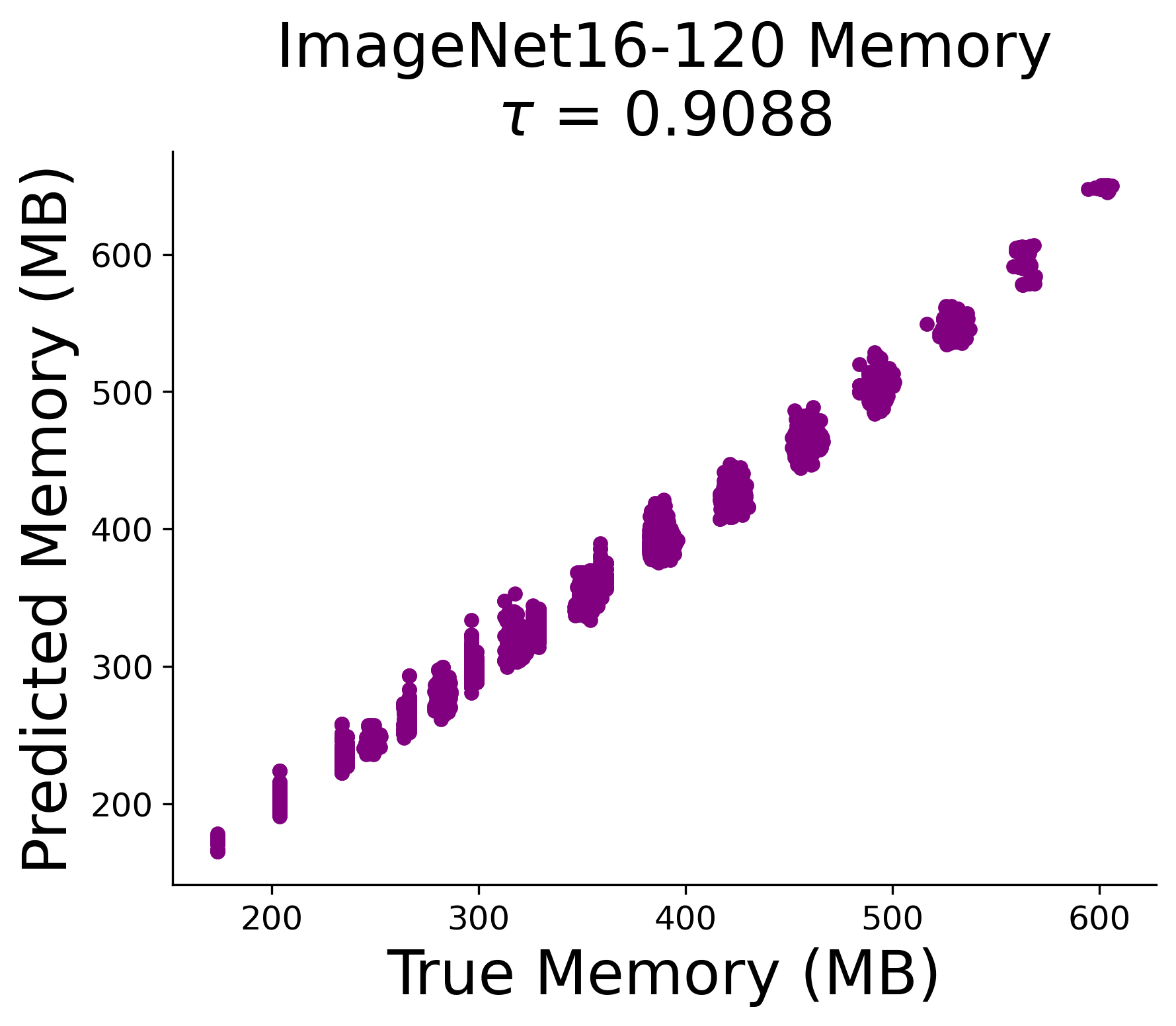}
            \caption{\scriptsize{ImageNet16-120}}
            \label{fig:subfig3}
        \end{subfigure}
    \end{minipage}

    \begin{minipage}{0.05\linewidth}
        \rotatebox{90}{\fontfamily{phv}\selectfont\large\textbf{\textcolor{orange}{Latency}}} 
    \end{minipage}
    \begin{minipage}{0.9\linewidth}
        \begin{subfigure}[b]{0.3\linewidth}
            \centering
            \includegraphics[width=\linewidth]{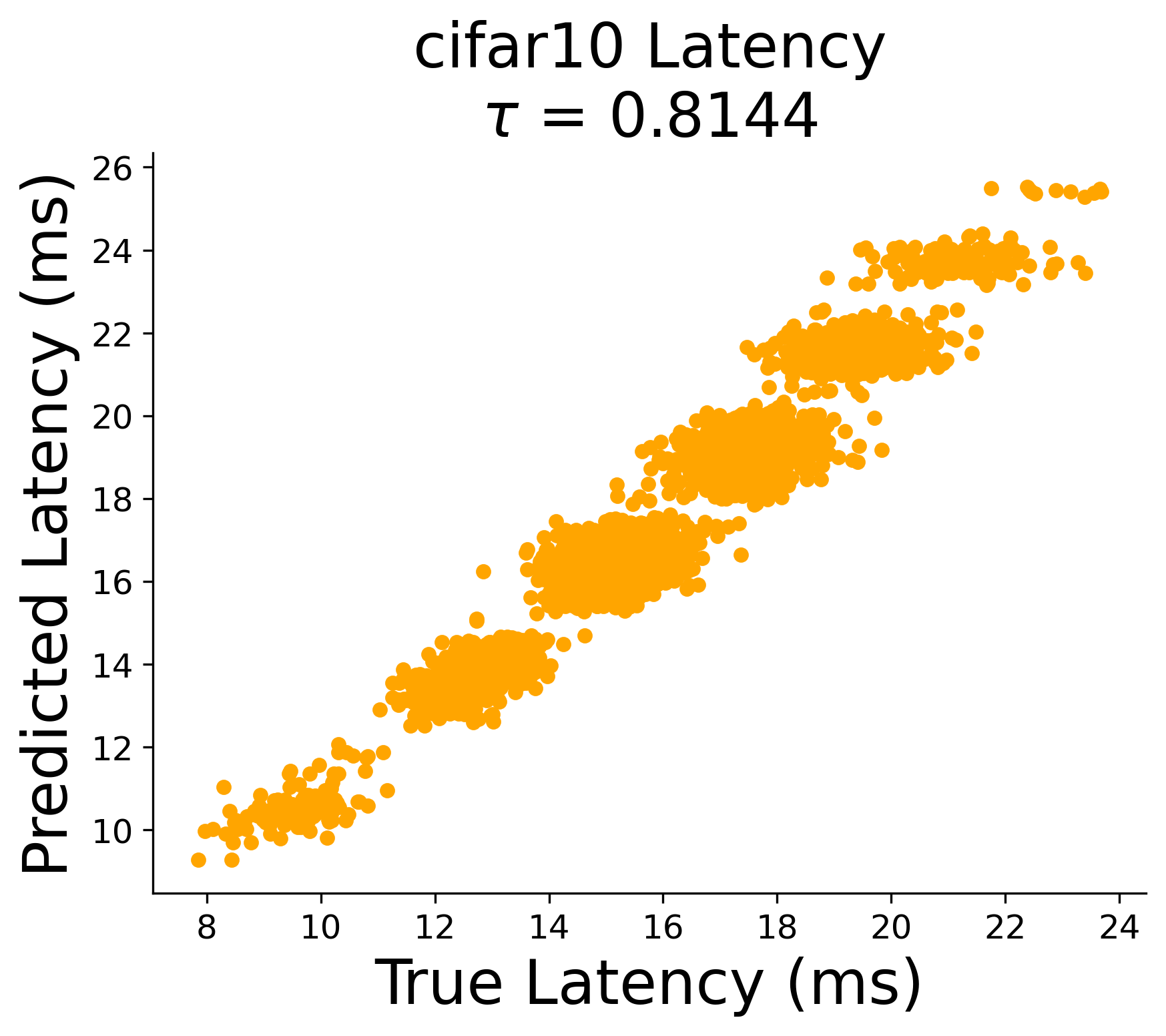}
            \caption{CIFAR-10}
            \label{fig:subfig4}
        \end{subfigure}
        \hfill
        \begin{subfigure}[b]{0.3\linewidth}
            \centering
            \includegraphics[width=\linewidth]{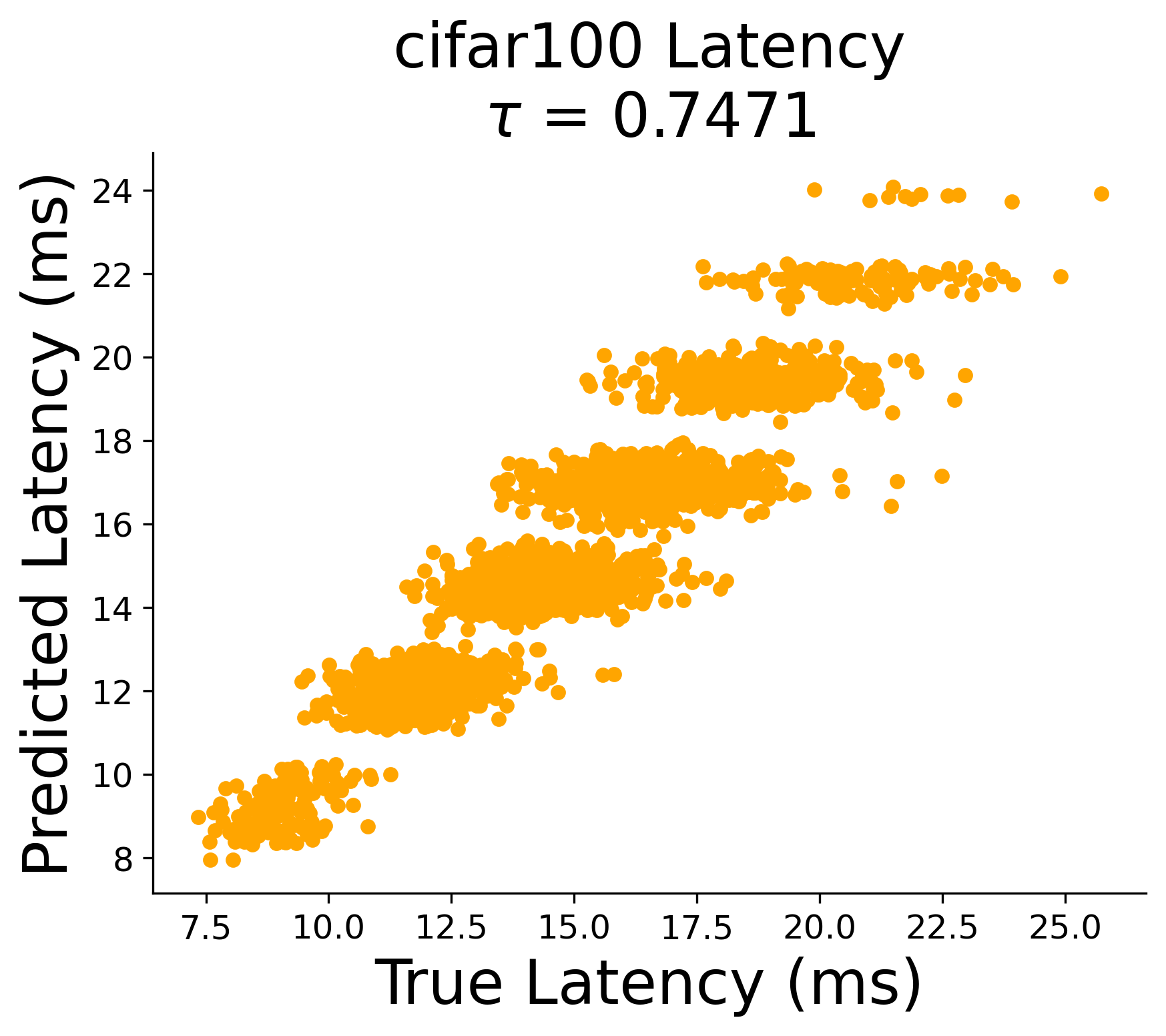}
            \caption{CIFAR-100}
            \label{fig:subfig5}
        \end{subfigure}
        \hfill
        \begin{subfigure}[b]{0.3\linewidth}
            \centering
            \includegraphics[width=\linewidth]{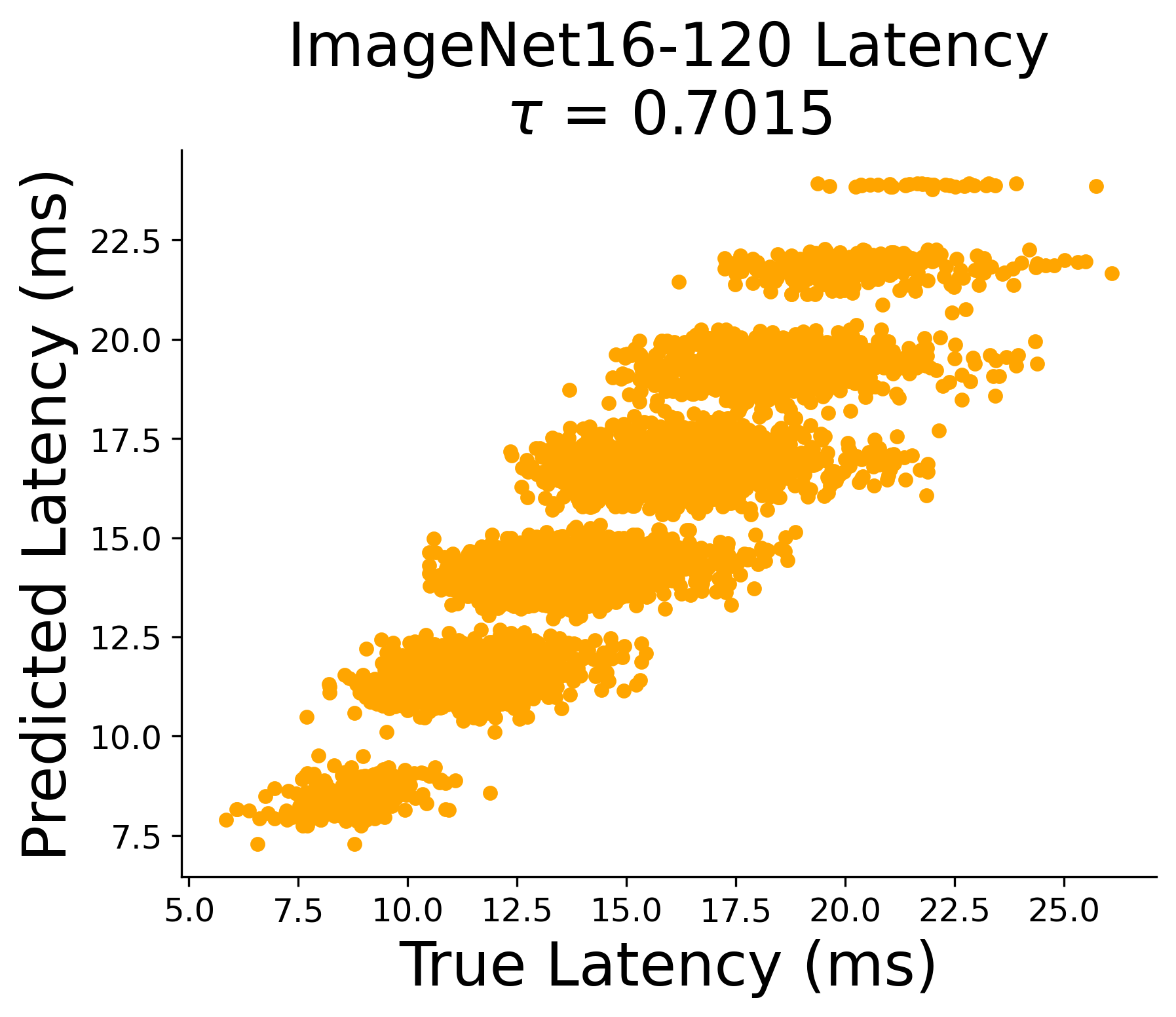}
            \caption{\scriptsize{ImageNet16-120}}
            \label{fig:subfig6}
        \end{subfigure}
    \end{minipage}

    \caption{Plots of predicted vs true hardware cost of \textcolor{blue}{NATS-Bench TSS architectures} (\textcolor{magenta}{T5-small}) for performance metrics reported on CIFAR-10, CIFAR-100, and ImageNet16-120. The strength of correlation increases as $\tau$ approaches 1.}
    \label{fig:plots_nats_tss}
\end{figure}

The results of the NATS-Bench TSS in Fig. \ref{fig:plots_nats_tss} show a strong positive Kendall $\tau$ correlation between predicted and true values for hardware costs. Predicted memory usage aligns closely with the true values across all datasets. Similarly, latency predictions exhibit a high correlation for CIFAR-10 and CIFAR-100, while ImageNet16-120 shows a slightly weaker correlation. This suggests that SEval-NAS effectively predicts hardware costs in the TSS space due to the architectural features. The reason why the SEval-NAS effectively predicts hardware costs in the TSS in Fig.~\ref{fig:plots_nats_tss} is due to the Autograd Traversal and String Generation block, which significantly optimizes the neural architecture in topology.

\begin{figure}[htbp]
    \centering
    \begin{minipage}{0.05\linewidth}
        \rotatebox{90}{\fontfamily{phv}\selectfont\large\textbf{\textcolor{violet}{Memory}}} 
    \end{minipage}\hspace{0pt}
    \begin{minipage}{0.92\linewidth}
        \begin{subfigure}[b]{0.3\linewidth}
            \centering
            \includegraphics[width=\linewidth]{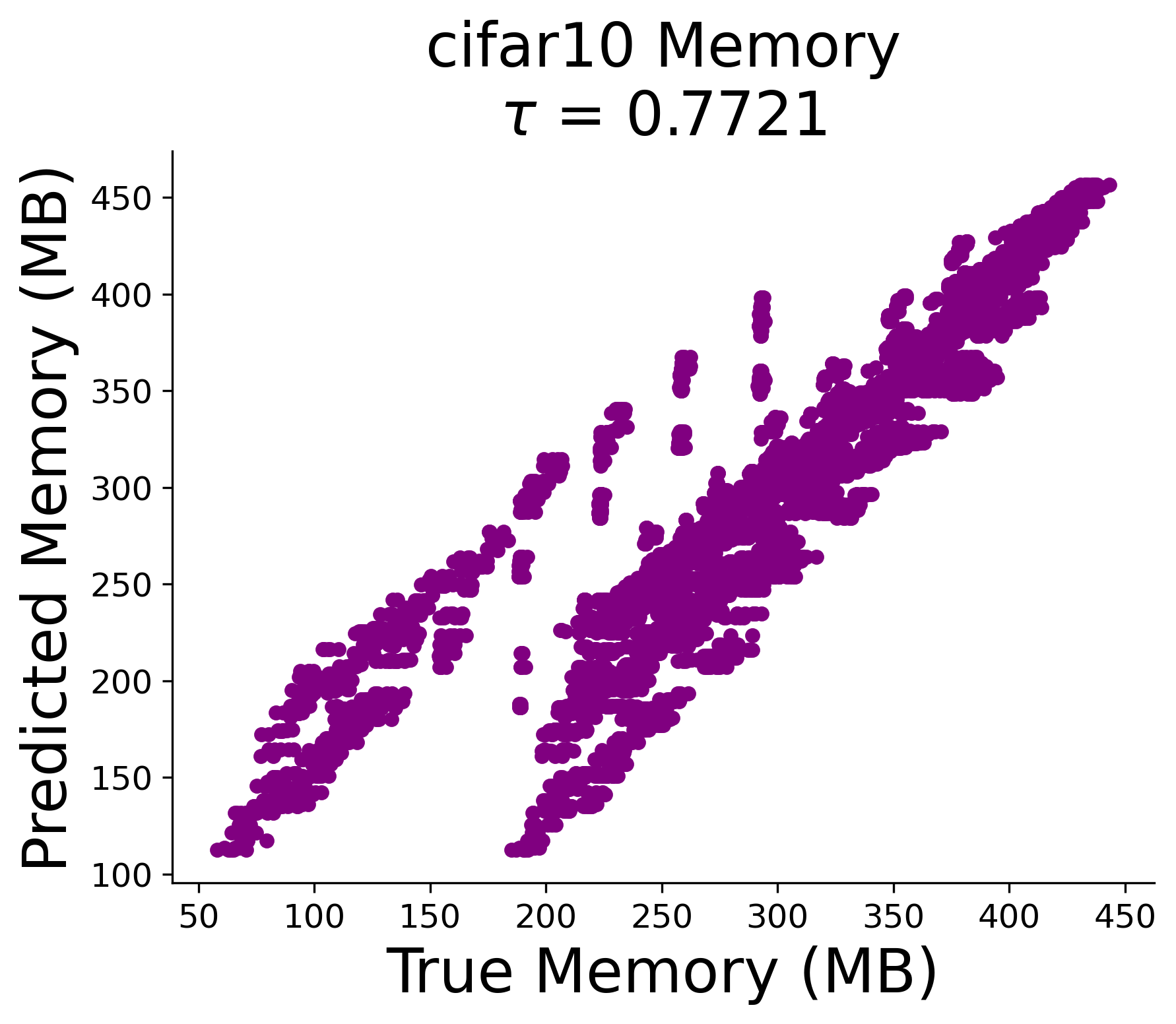}
            \caption{CIFAR-10}
        \end{subfigure}
        \hfill
        \begin{subfigure}[b]{0.3\linewidth}
            \centering
            \includegraphics[width=\linewidth]{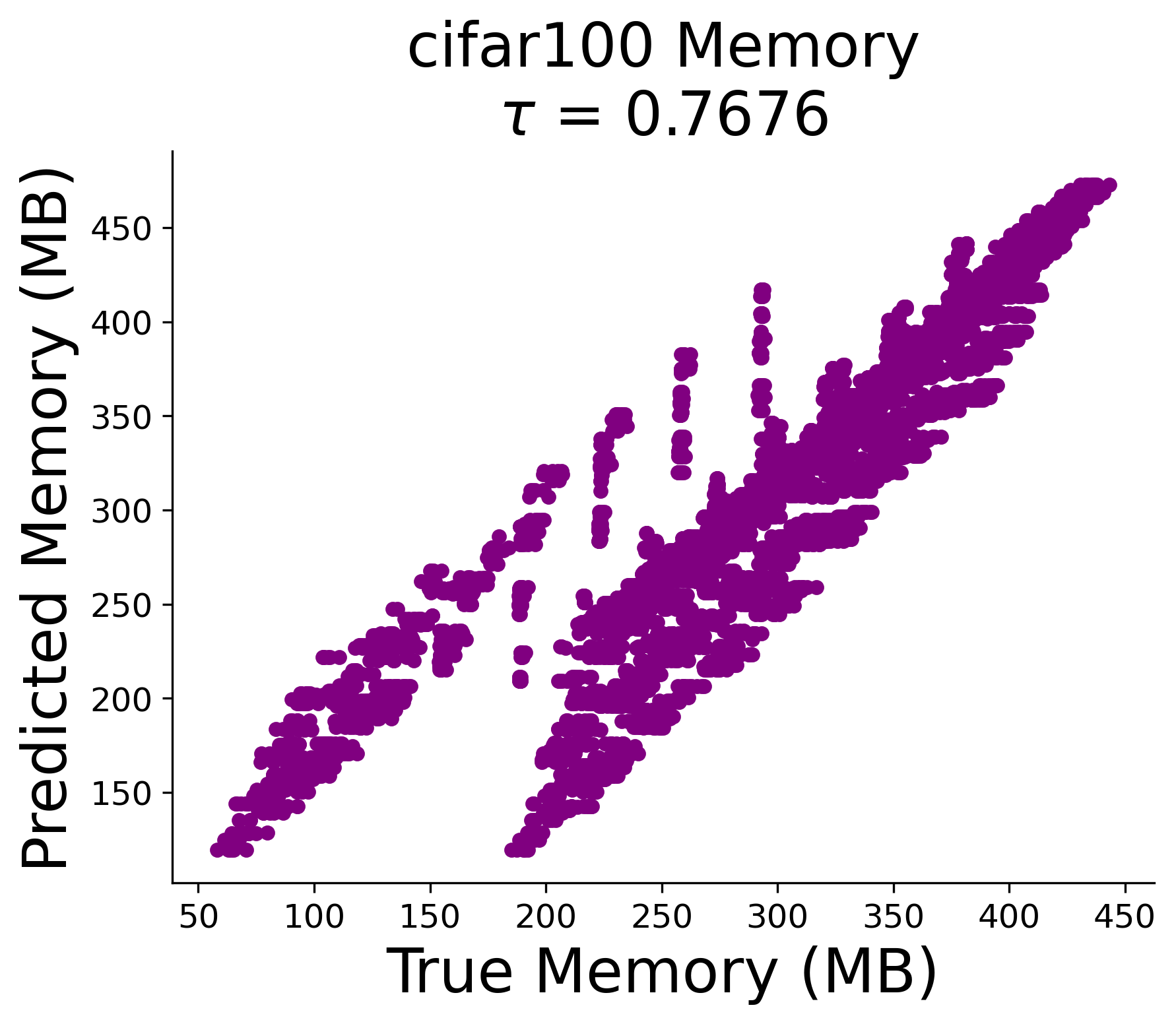}
            \caption{CIFAR-100}
        \end{subfigure}
        \hfill
        \begin{subfigure}[b]{0.3\linewidth}
            \centering
            \includegraphics[width=\linewidth]{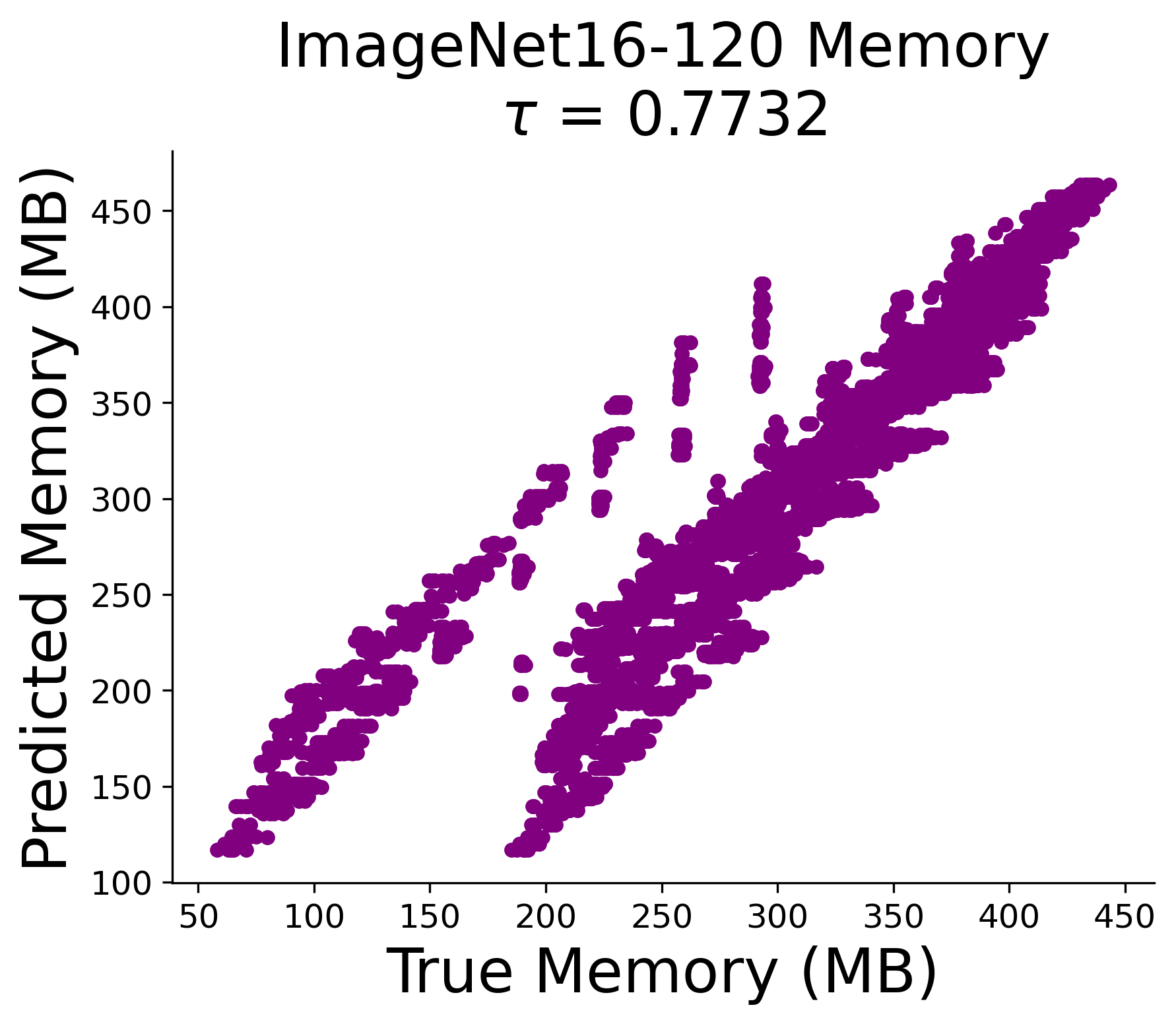}
            \caption{\scriptsize{ImageNet16-120}}
        \end{subfigure}
    \end{minipage}

    \begin{minipage}{0.05\linewidth}
        \rotatebox{90}{\fontfamily{phv}\selectfont\large\textbf{\textcolor{orange}{Latency}}} 
    \end{minipage}\hspace{0pt}
    \begin{minipage}{0.92\linewidth}  
        \begin{subfigure}[b]{0.3\linewidth}
            \centering
            \includegraphics[width=\linewidth]{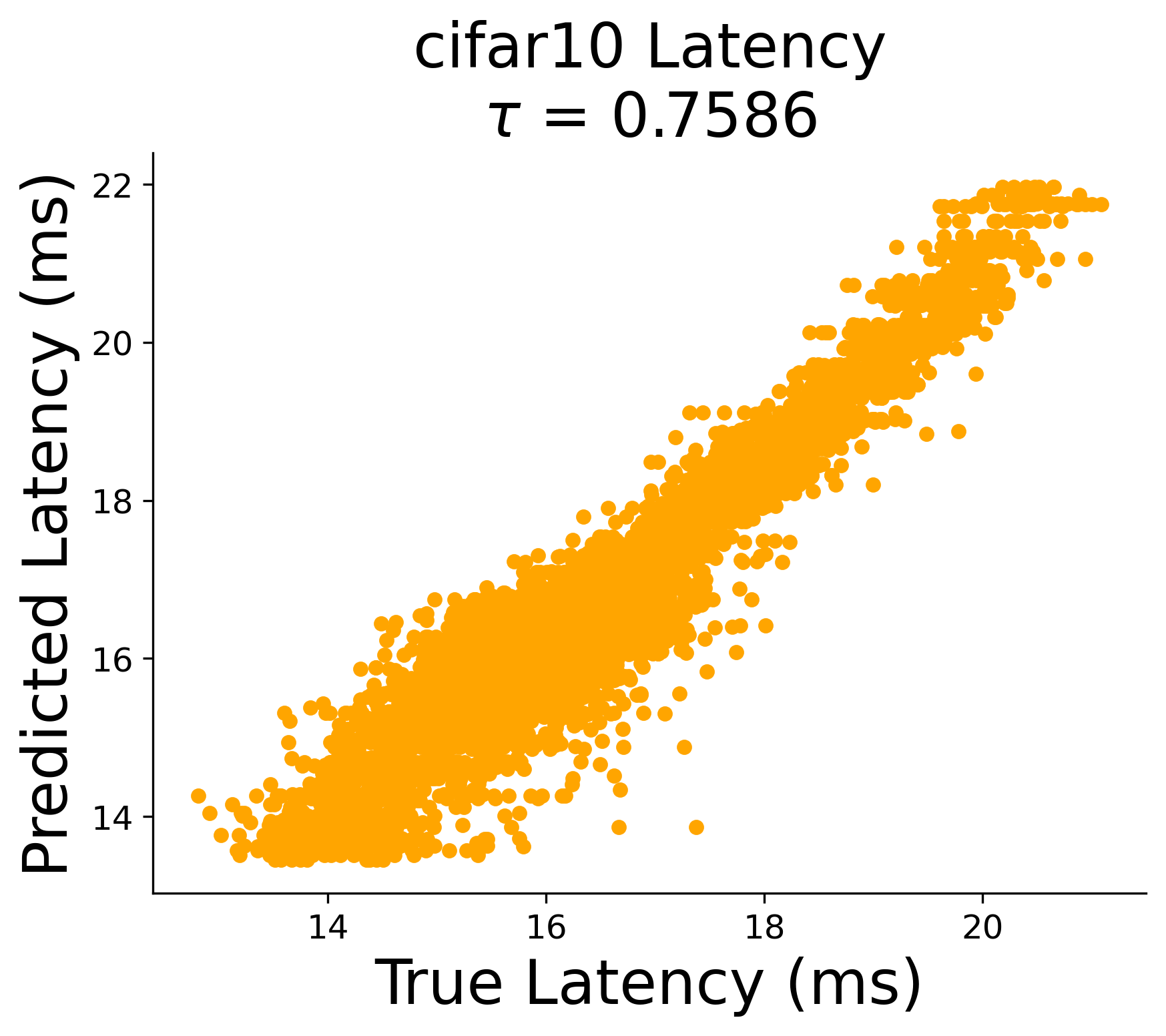}
            \caption{CIFAR-10}
        \end{subfigure}
        \hfill
        \begin{subfigure}[b]{0.3\linewidth}
            \centering
            \includegraphics[width=\linewidth]{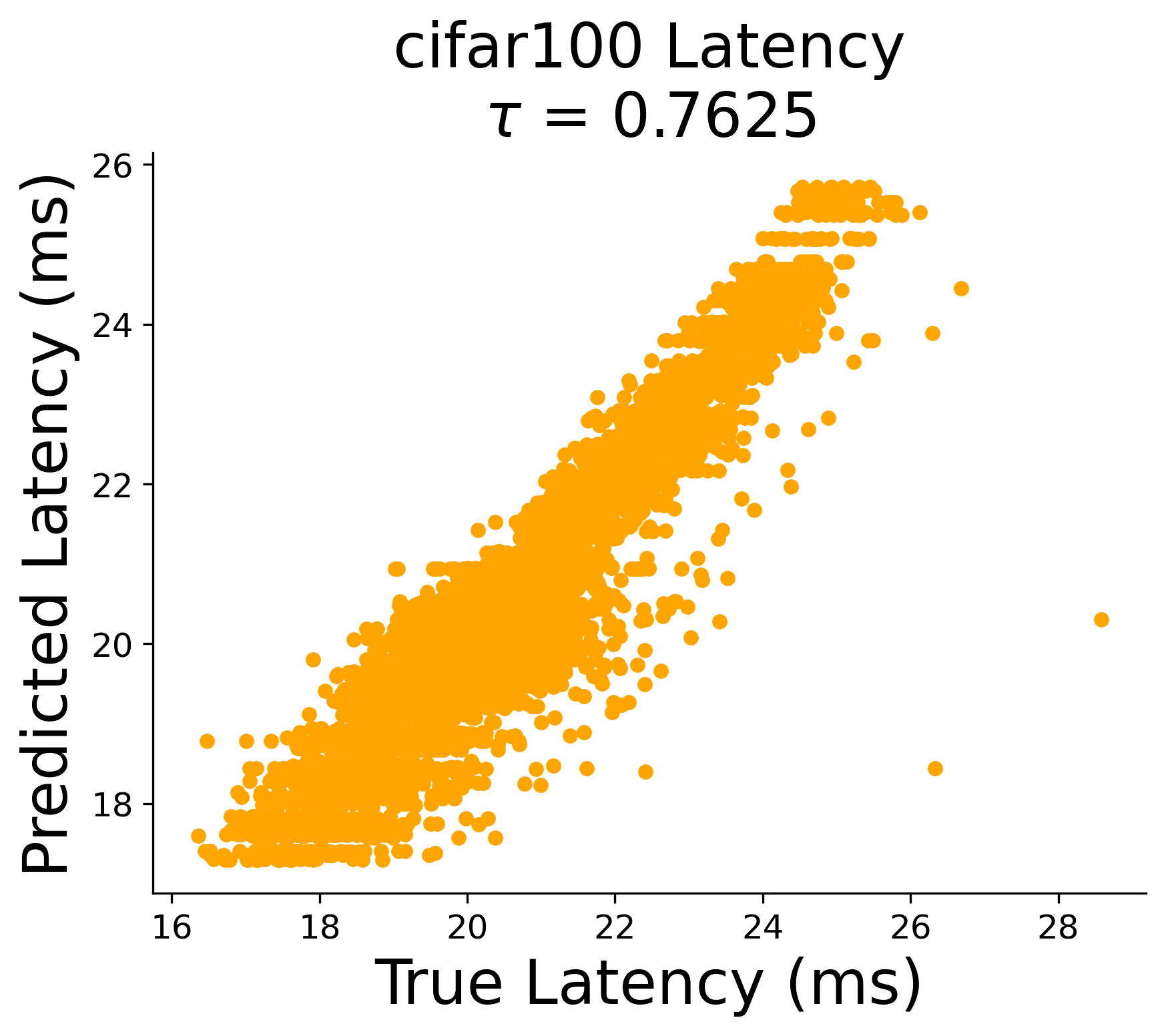}
            \caption{CIFAR-100}
        \end{subfigure}
        \hfill
        \begin{subfigure}[b]{0.3\linewidth}
            \centering
            \includegraphics[width=\linewidth]{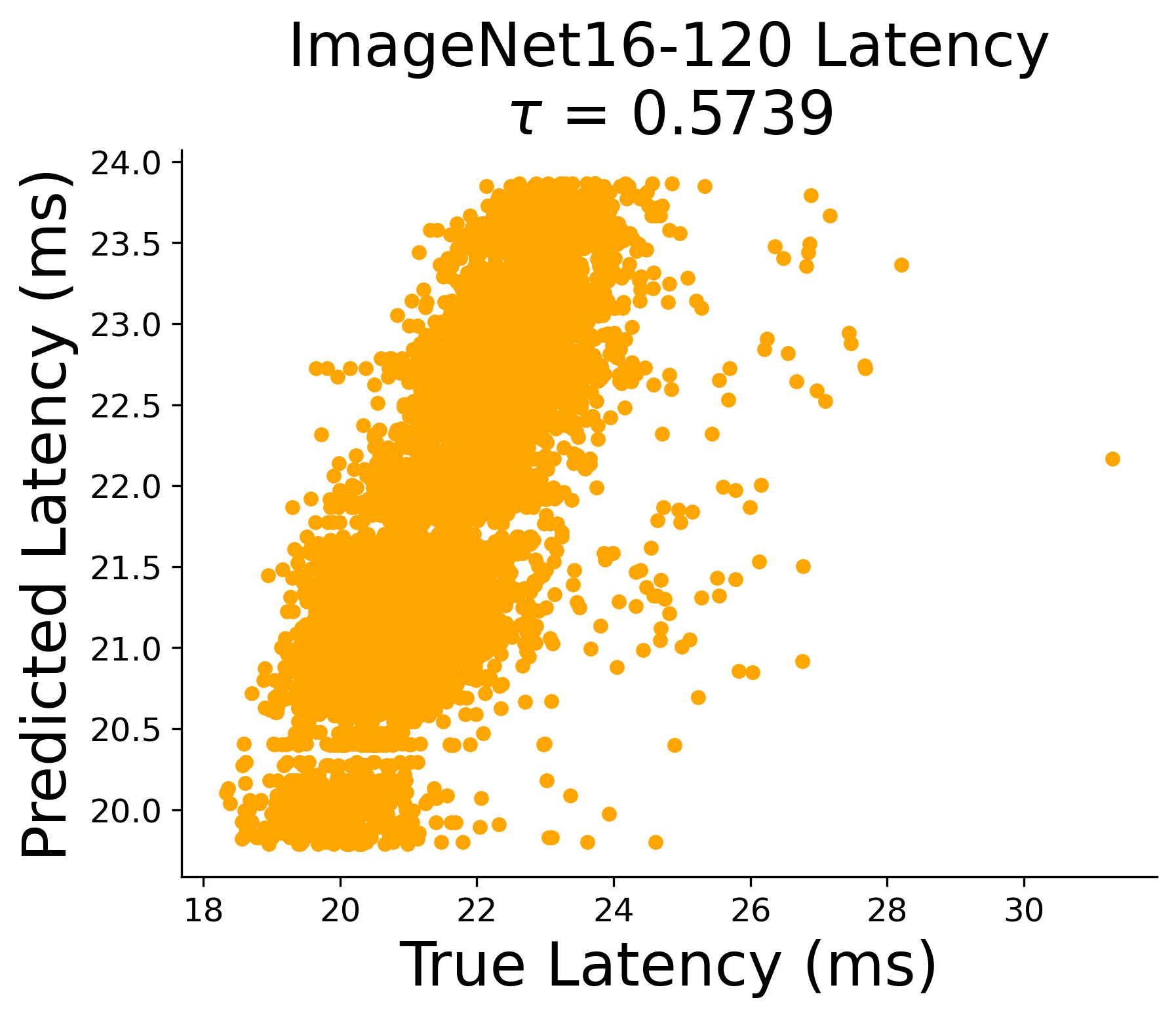}
            \caption{\scriptsize{ImageNet16-120}}
        \end{subfigure}
    \end{minipage}
    \caption{Plots of predicted vs true hardware cost of \textcolor{green}{NATS-Bench SSS architectures} (\textcolor{magenta}{T5-small}) for performance metrics reported on CIFAR-10, CIFAR-100, and ImageNet16-120. The strength of correlation increases as $\tau$ approaches 1.}
    \label{fig:plots_nats_sss}
\end{figure}

For the NATS-Bench SSS results illustrated in Fig. \ref{fig:plots_nats_sss}, predicted memory usage again shows a strong positive correlation with the true values across all datasets. Predicted latency has a similar trend with a strong correlation in CIFAR-10 and CIFAR-100. However, latency prediction for the ImageNet16-120 dataset appears less robust, with noticeable variance. In terms of predicted latency on SSS across CIFAR-10, CIFAR-100, and ImageNet16-120 in Fig.~\ref{fig:plots_nats_tss}. From the dataset itself, especially ImageNet16-120 compared to CIFAR-100 and CIFAR-10, we know ImageNet16-200 emphasizes low-resolution feature extraction; thus, ImageNet16-120 is reliable for memory predictions but unstable for latency prediction. Whereas CIFAR-100 and CIFAR-10 are designed for fine-grained classification, which results in good reliability on both memory predictions and latency memory. Therefore, while SEval-NAS remains reliable for memory predictions, its reliability in predicting latency is dataset-dependent and influenced by the variability of the architectures.

\begin{figure}[h]
    \centering
    \begin{subfigure}[b]{0.49\linewidth}
        \centering
        \includegraphics[width=\linewidth]{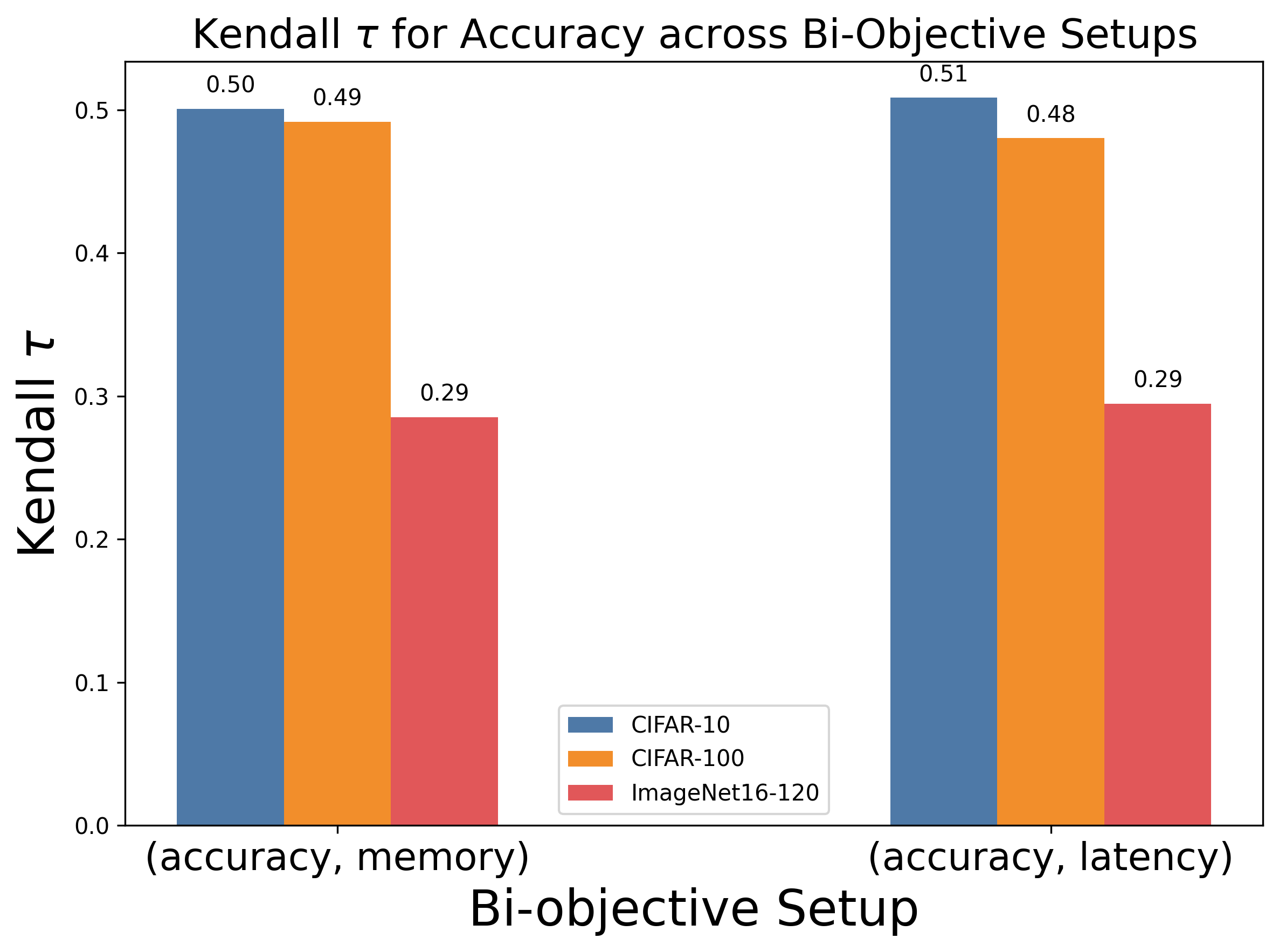}
        \caption{\textcolor{blue}{NATS-Bench TSS}}
    \end{subfigure}
    \begin{subfigure}[b]{0.49\linewidth}
        \centering
        \includegraphics[width=\linewidth]{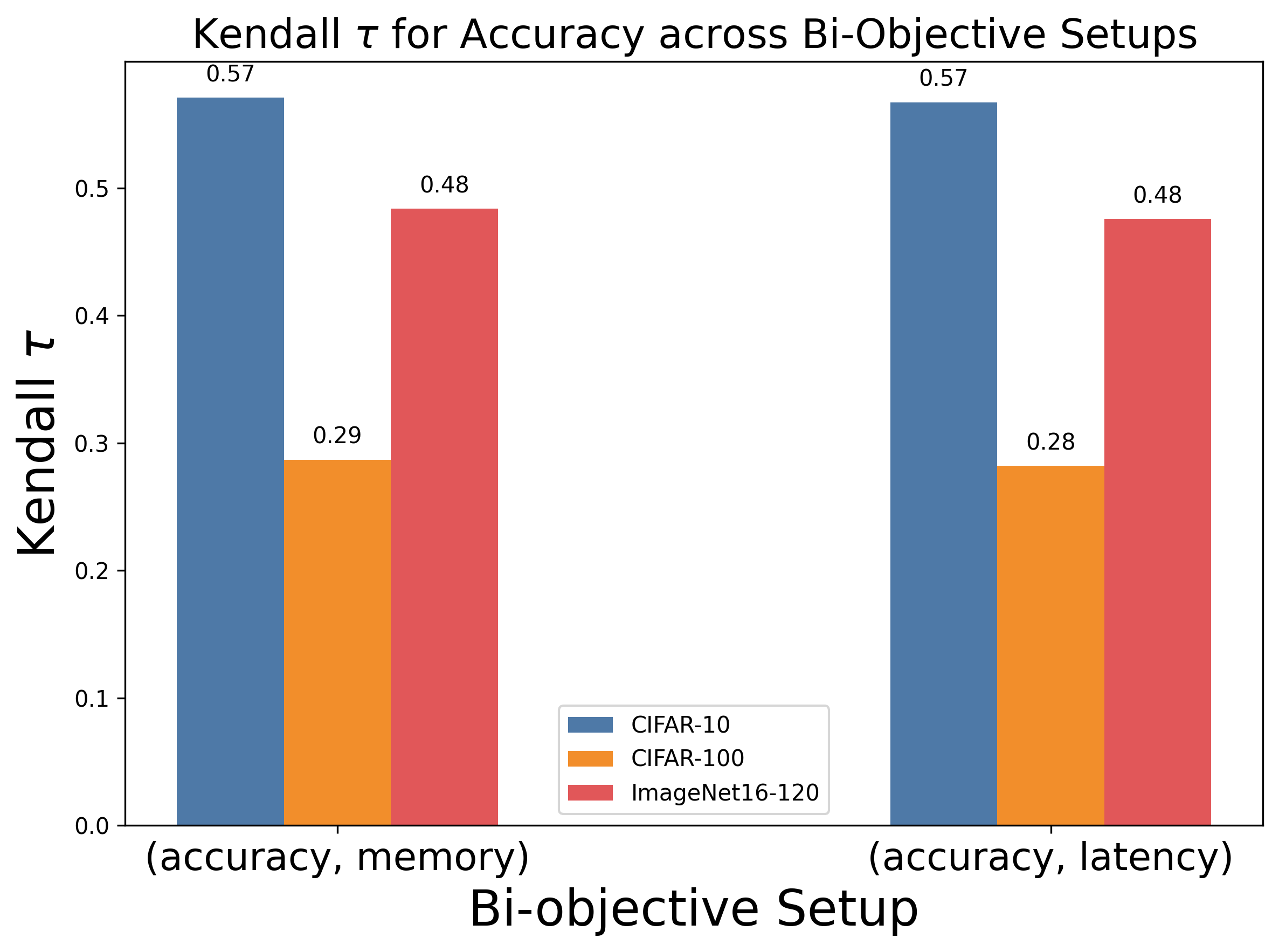}
        \caption{\textcolor{green}{NATS-Bench SSS}}
    \end{subfigure}
    \caption{Kendall's $\tau$ correlation for predicted vs true accuracy in NATS-Bench TSS and SSS search spaces. Comparison includes \textit{(accuracy, latency)} and \textit{(accuracy, memory)} bi-objectives.}
    \label{fig:graph_accuracy}
\end{figure}

In comparison, while positively correlated, accuracy predictions in Fig.~\ref{fig:graph_accuracy} exhibit weaker correlations than those for hardware costs. This suggests that SEval-NAS struggles to confidently infer accuracy from neural architecture representations. Furthermore, there is no trend linking dataset type to prediction reliability for accuracy, highlighting that accuracy depends on factors beyond straightforward architectural features. Overall, SEval-NAS demonstrates stronger predictability for hardware metrics than accuracy, primarily because hardware costs are directly tied to architectural characteristics. For example, more convolutional filters in a network will need more computation than fewer convolutional filters \cite{Mih2024}.
Meanwhile, this correlation cannot be directly made for accuracy.


We also conducted an experiment (Appendix \ref{appd:A}) to evaluate three different encoder/decoder models (T5-small, T5-base, and T5-large) on NATS-Bench SSS and NATS-Bench TSS. This ablation study compared how model size affects the correlation between predicted versus true memory and predicted versus true latency, as measured by Kendall correlation.

The results from our extra ablation studies (Fig. \ref{fig:encoder_kendall} in Appendix \ref{appd:A.1}) showed that T5-small, T5-base, and T5-large perform similarly on the TSS benchmark. However, we observed that only T5-large encoders exhibit lower Kendall $\tau$ correlations on the SSS.
Additionally, we found that different encoder sizes (Appendix \ref{appd:A.2}) do not significantly impact performance on NATS-Bench TSS for either memory or latency correlations (Fig. \ref{fig:encoders_tss}). In contrast, T5-large demonstrates weaker $\tau$ correlations for both memory and latency predictions on NATS-Bench SSS, as shown in Fig. \ref{fig:encoders_sss}.


\subsection{Experiment 2: Predicting Hardware Cost (Evaluation on HW-NAS-Bench)}
Although we observe a strong positive latency correlation in the NATS-Bench search spaces, the latency of a NN largely depends on the hardware environment. To investigate how well SEval-NAS would theoretically predict latency across various hardware devices, we evaluate its performance on HW-NAS-Bench. 

HW-NAS-Bench \cite{LiC2021} was designed for hardware-aware NAS. It includes two NAS search space designs: NAS-Bench-201's cell-based search space and FBNet's search space. The dataset provides the hardware cost of the NNs from both search spaces on commercial devices, including Edge GPU, Edge TPU, ASIC Eyeriss, FPGA, Pixel 3, and Raspberry Pi 4. 
FBNet search space \cite{WuB2019} builds a layer-wise search space with a fixed macro-architecture and varying middle layers that can be searched. The architectures in this search space have regular structures that include nine cell candidates and 22 positions, yielding $9^{22} \approx 10^{21}$ different architectures.
Due to the excessively large size of the search space, we do not use it in our experiment. 
NAS-Bench-201 search space \cite{Dong2020} is the original search space of the TSS architectures in NATS-Bench. It contains the same 15,625 architectures with results reported for CIFAR-10, CIFAR-100, and ImageNet16-120 and their hardware costs on each of the six devices. 
We evaluate SEval-NAS on the HW-NAS-Bench's NAS-Bench-201 subspace for values reported on the CIFAR-10 dataset. The evaluator is trained to predict only latency, testing the performance of SEval-NAS on a single objective metric.



\begin{figure}[htbp]
    \centering
    \begin{subfigure}[b]{0.32\linewidth}
        \centering
        \includegraphics[width=\linewidth]{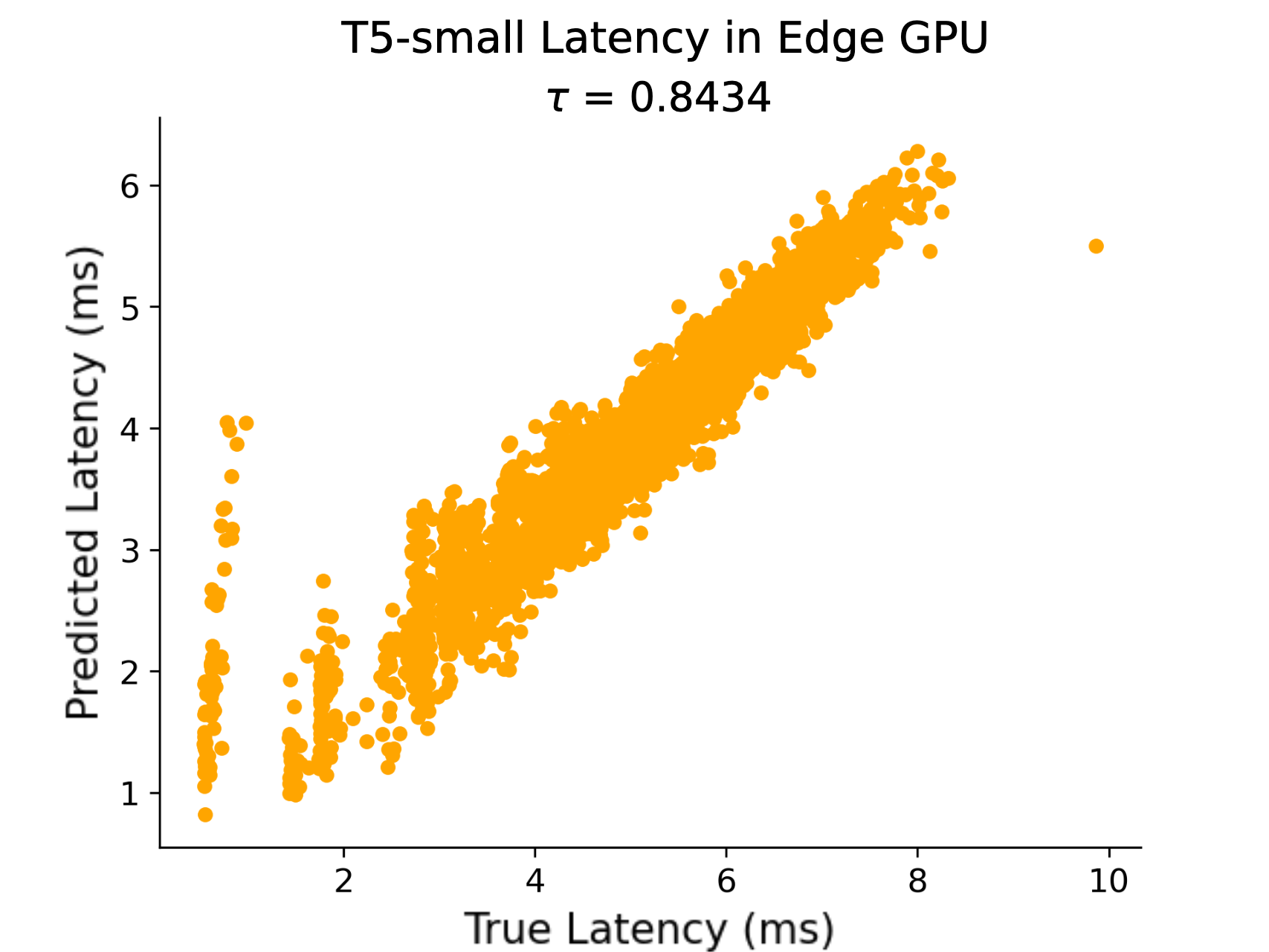}
        \caption{Edge GPU}
    \end{subfigure}
    \hfill
    \begin{subfigure}[b]{0.32\linewidth}
        \centering
        \includegraphics[width=\linewidth]{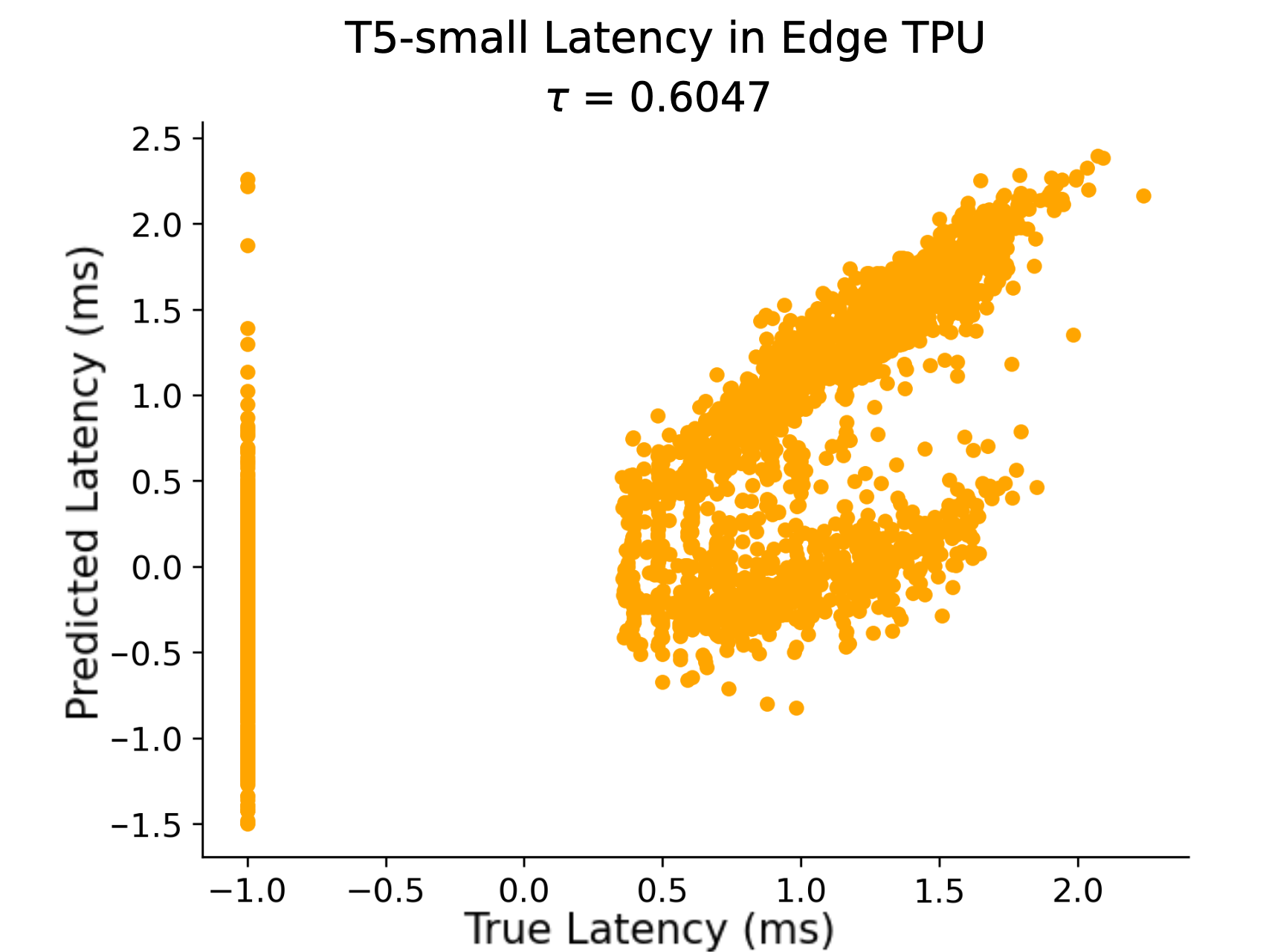}
        \caption{Edge TPU}
    \end{subfigure}
    \hfill
    \begin{subfigure}[b]{0.32\linewidth}
        \centering
        \includegraphics[width=\linewidth]{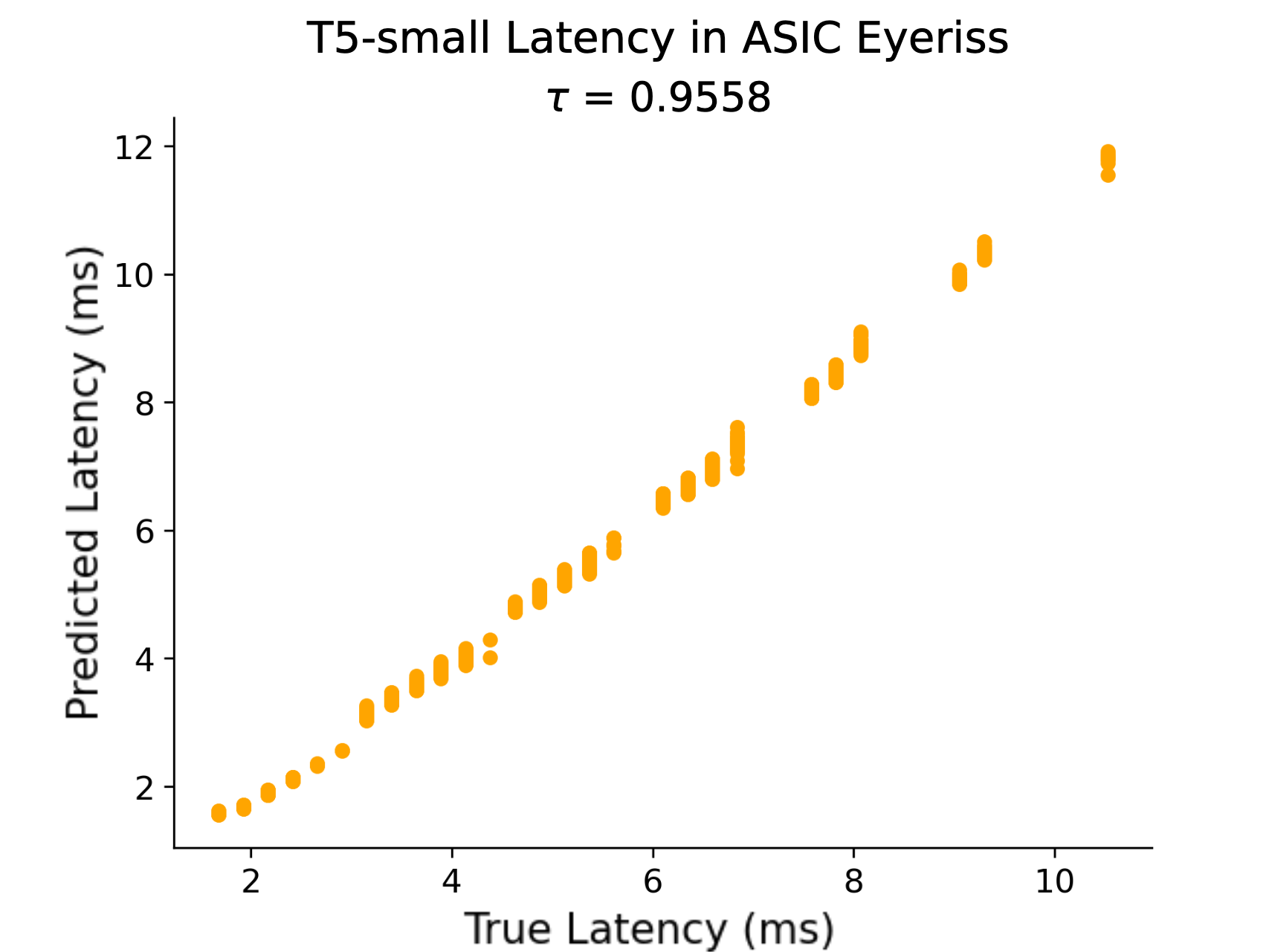}
        \caption{ASIC Eyeriss}
    \end{subfigure}
    
    \begin{subfigure}[b]{0.32\linewidth}
        \centering
        \includegraphics[width=\linewidth]{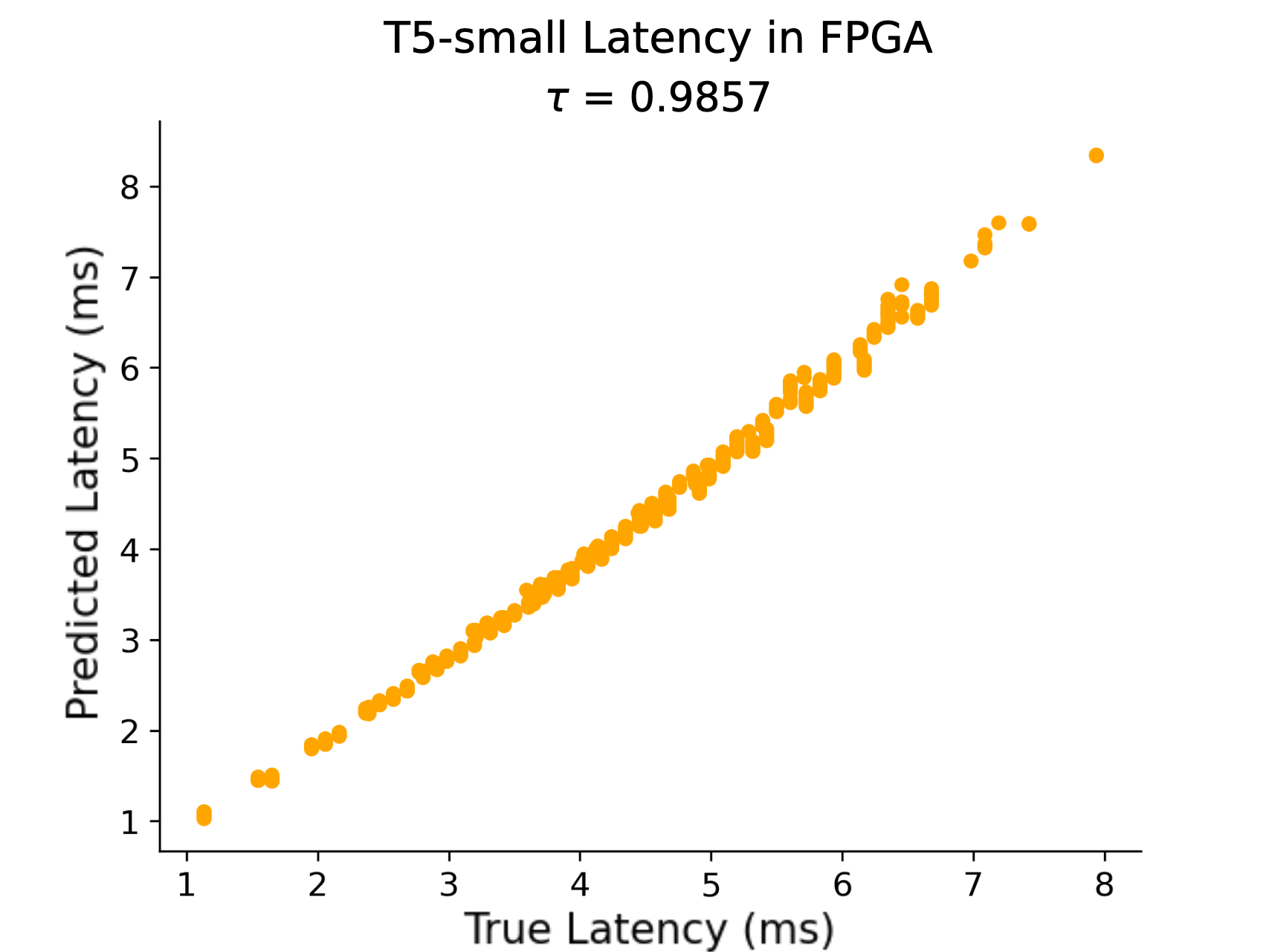}
        \caption{FPGA}
    \end{subfigure}
    \hfill
    \begin{subfigure}[b]{0.32\linewidth}
        \centering
        \includegraphics[width=\linewidth]{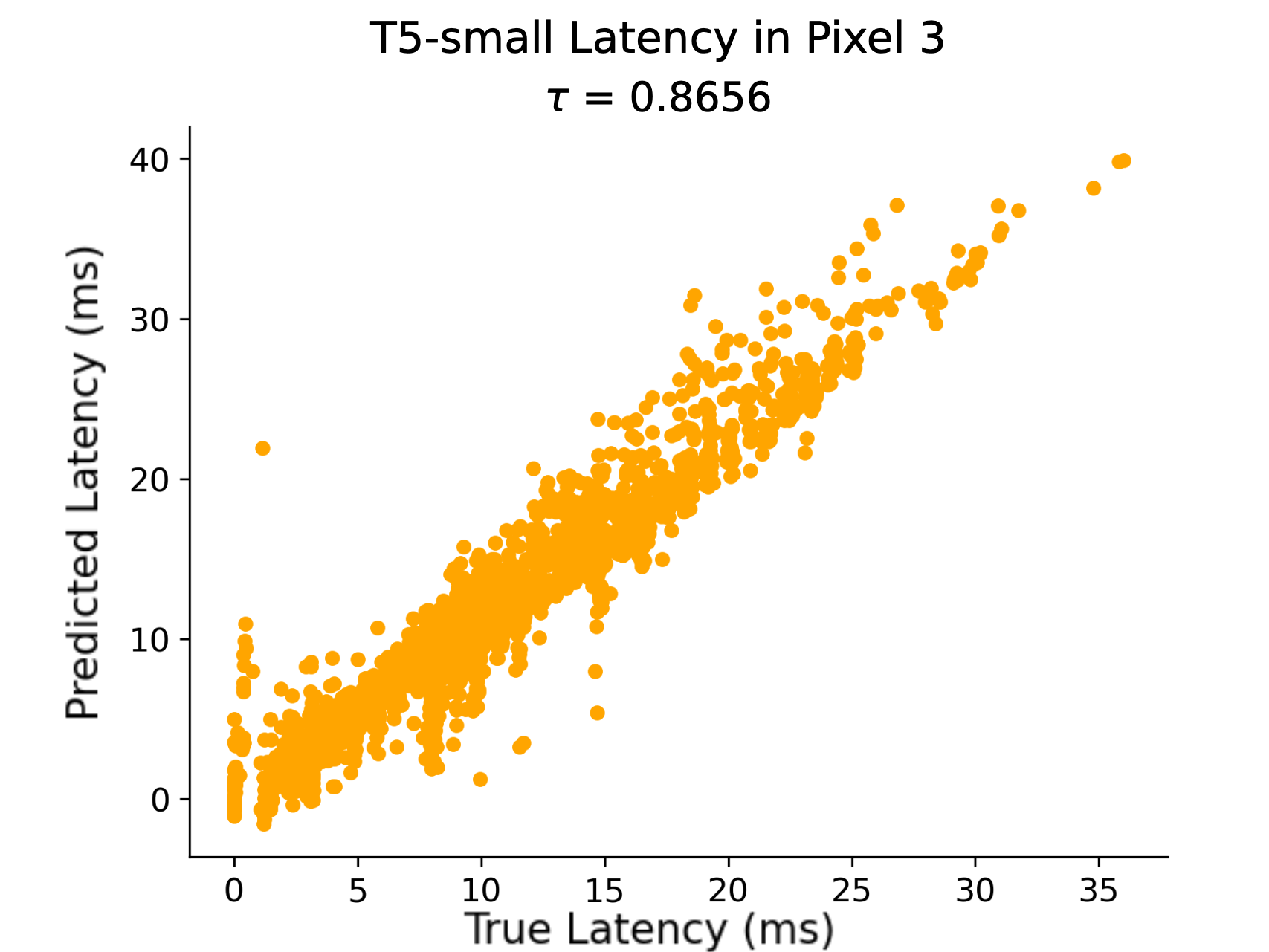}
        \caption{Pixel 3}
    \end{subfigure}
    \hfill
    \begin{subfigure}[b]{0.32\linewidth}
        \centering
        \includegraphics[width=\linewidth]{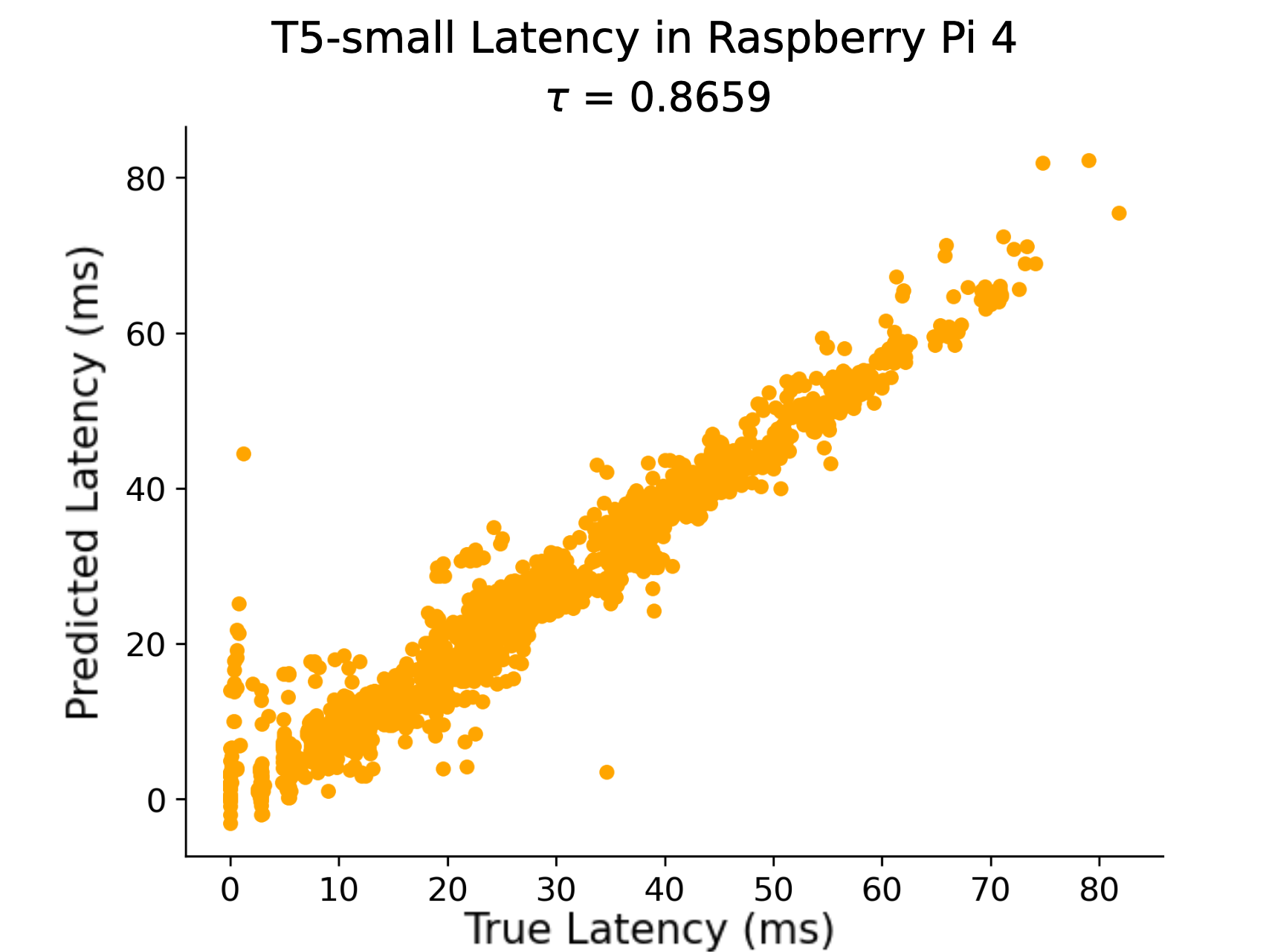}
        \caption{Raspberry Pi 4}
    \end{subfigure}
    \caption{Plots of predicted vs true latency for NAS-Bench-201 architectures with \textcolor{magenta}{T5-small} encoder for 6 edge devices reported in the HW-NAS-Bench benchmark. The strength of correlation increases as $\tau$ approaches 1.}
    \label{fig:plots_hwnas_small}
\end{figure}

Results in Fig. \ref{fig:plots_hwnas_small} show a strong positive Kendall $\tau$ correlation for most edge devices, with values ranging from 0.6047 to 0.9742, demonstrating SEval-NAS's ability to predict latency across six different edge devices. The Edge TPU's latency predictions stand out as an outlier, showing a weaker correlation ($\tau = 0.6047$). This is attributed to negative latency values reported in the HW-NAS-Bench dataset, which likely affected the model's ability to predict latency in this case. Despite this, other devices, such as the Edge GPU ($\tau = 0.8676$), Eyeriss ($\tau = 0.9558$), FPGA ($\tau = 0.9742$), Pixel 3 ($\tau = 0.8599$), and Raspi4 ($\tau = 0.8659$), exhibit strong correlations, indicating consistent and reliable performance across diverse hardware configurations.

Both experiments show that training the evaluator for bi-objectives in NATS-Bench and a single objective in HW-NAS-Bench consistently yielded positive Kendall $\tau$ correlation values. This demonstrates the effectiveness of SEval-NAS in adapting to different numbers of evaluation objectives, further solidifying its strength as a predictive method for latency across edge devices.

We also conduct the additional experiments (in Appendix \ref{appd:A.3}) to run three different encoder/decoder models: T5-small (Fig. \ref{fig:plots_hwnas_small}), T5-base (Fig. \ref{fig:plots_hwnas_base}), and T5-large (Fig. \ref{fig:plots_hwnas_large}) on HW-NAS-Bench as an ablation study of comparison of model size on the predicted latency vs true latency. 

We observe that different encoder/decoder models do not have great impact (less than 0.02 latency correlation difference) except that T5-base in Edge GPU (0.8804) and T5-large in Edge GPU (0.8852) have stronger latency correlation than T5-small in Edge GPU(0.8434) since T5-small's operators are smaller, which results in GPU kernel launching overhead accounting for a higher proportion of the total latency. Moreover, T5-large and T5-base have longer latency, the relative impact of noise is smaller, which leads to a higher latency correlation.





\subsection{Experiment 3: Ease of Integration (Evaluator in a NAS)}
The contrast in correlation values observed between accuracy (low to moderate correlation) and hardware costs (strong correlation) highlights the evaluator's strength as a hardware cost predictor. This strength can be effectively leveraged to enhance NAS approaches that traditionally optimize accuracy as a single objective by incorporating hardware constraints into their search strategies. To demonstrate this applicability, we integrate SEval-NAS into FreeREA \cite{Cavagnero2023} to search in NATS-Bench search space and define hardware constraints for the search. FreeREA uses evolutionary search to find candidate architectures and evaluates candidate architectures using a training-free metric to estimate the accuracy performance. 

FreeREA algorithm on NATS-Bench TSS is constrained by $FLOPS$ and $\#Params$. Since FLOPS is a poor proxy for hardware costs \cite{LiC2021, Sinha2024}, we replace it with two alternative constraints: 1) latency and 2) memory usage, each tested separately.
For each metric, the mean value reported in the benchmark is set as the threshold for ranking candidate architectures. For example, these cases use the mean latency (45.96 seconds) and memory usage (166.67 MB) from the NATS-Bench CIFAR-10 dataset as thresholds. The resulting performance is compared against FreeREA and other training-free NAS algorithms reported in \cite{Cavagnero2023}, with findings summarized in Table \ref{tab:nas_compare}.

\begin{table}[]
    \centering
    \caption{Test accuracy and time for various NAS algorithms for NATS-Bench.}
    \renewcommand{\arraystretch}{1.2} 

\resizebox{\linewidth}{!}{
\rowcolors{2}{white}{gray!10}
\begin{tabular}{l|lr|lr|lr}
\rowcolor{cyan!40}
\hline
\multicolumn{1}{l|}{}          & \multicolumn{2}{c|}{\textbf{CIFAR 10}}           & \multicolumn{2}{c|}{\textbf{CIFAR 100}}          & \multicolumn{2}{c}{\textbf{ImageNet16-120}} \\ \hline
\rowcolor{gray!30}
\multicolumn{1}{l|}{Algorithm} & Accuracy    & \multicolumn{1}{l|}{Time (s)} & Accuracy    & \multicolumn{1}{l|}{Time (s)} & Accuracy            & Time (s)        \\ \hline
NASWOT (1000)                  & $93.10 \pm 0.31$  & $248$                       & $69.10 \pm 1.61$  & $248$                       & $45.08 \pm 1.55$          & $248$          \\
TENAS                          & $93.90 \pm 0.47$   & $1558$                      & $71.24 \pm 0.56$   & $1558$                      & $42.38 \pm 0.46$          & $1558$         \\
NASI                           & $93.55 \pm 0.10$  & $120$                       & $71.20 \pm 0.14$  & $120$                       & $44.84 \pm 1.41$          & $120$          \\
GA-NINASWOT                    & $93.70 \pm 0.63$  & $206$                       & $71.57 \pm 1.37$  & $206$                       & $45.18 \pm 2.05$          & $206$          \\
EPE-NAS                        & $91.31 \pm 1.69$  & $104$                       & $69.58 \pm 0.83$  & $104$                       & $41.84 \pm 2.06$          & $104$          \\
\rowcolor{green!30}
FreeREA                        & $94.36 \pm 0.00$  & $45$                        & $73.51 \pm 0.05$  & $45$                        & $46.34 \pm 0.00$          & $45$           \\ \hline
\rowcolor{pink!20}
\tikzmark{tl}FreeREA + Latency              & $94.36 \pm 0.00$  & $77$                      & $73.51 \pm 0.00$  & $82$                        & $46.34 \pm 0.00$          & $81$           \\
\rowcolor{pink!30}
FreeREA + Memory               & $84.21 \pm 14.91$ & $34$                        & $49.67 \pm 11.35$ & $37$                        & $19.66 \pm 7.19$          & $38$\tikzmark{br}           \\ \hline
\end{tabular}
}
\begin{tikzpicture}[remember picture,overlay]
\draw[red,dashed,thick] ([xshift=-6pt,yshift=23pt]pic cs:tl) rectangle ([xshift=-131pt,yshift=19pt]pic cs:br);
\end{tikzpicture}
\vspace{-3em}
    \label{tab:nas_compare}
\end{table}



Latency-constrained search identified an average of approximately 230 architectures satisfying the threshold, achieving final average accuracies consistent with those reported in the original FreeREA study. In contrast, the memory-constrained search discovered fewer architectures (on average, fewer than 10) that met the threshold, indicating a bias in the search algorithm against low-memory architectures. This smaller pool of candidates led to higher variability in test accuracies due to the diverse performance of low-memory architectures. Importantly, while the latency-constrained search doubled FreeREA's search time, this overhead from evaluator inference remained negligible compared to other NAS algorithms. Conversely, the memory-constrained search required less time due to the limited number of viable candidate architectures.

These results highlight the flexibility of SEval-NAS for hardware cost evaluation while making minimal changes to the algorithm. By integrating additional constraints (e.g., latency and memory thresholds), the NAS algorithm can be tailored to select candidate architectures suitable for target hardware devices. For instance, deploying SEval-NAS on an edge device with memory constraints matching the device's operating range would yield architectures suitable for deployment.
This study establishes the feasibility of incorporating SEval-NAS into NAS pipelines, with the specific objective of demonstrating integration viability rather than algorithmic optimization. The investigation of threshold parameter effects on search dynamics represents a natural extension of this foundational work and constitutes a promising direction for future research.

\section{Conclusion and Future Work} \label{sec:conclusion}
NAS discovers novel architectures without expert knowledge but incurs long evaluation times when training or deploying candidate models. We proposed SEval-NAS, which converts architectures to string representations via autograd graph traversal and maps their embeddings to predicted performance metrics.
We evaluated SEval-NAS on NATS-Bench and HW-NAS-Bench for accuracy, latency, and memory. Latency and memory predictions showed the strongest correlations, demonstrating SEval-NAS as an effective hardware cost predictor, while accuracy predictions showed moderate correlation and revealed limitations in estimating accuracy.
Ablation studies across encoder and decoder sizes showed that larger encoders reduced Kendall $\tau$ correlations on NATS-Bench SSS. On HW-NAS-Bench, larger models improved latency correlation on Edge GPU due to kernel operator effects, but also increased latency.
To assess adaptability, we integrated SEval-NAS into FreeREA \cite{Cavagnero2023} by adding latency and memory constraints. SEval-NAS introduced minimal search-time overhead and enabled new evaluation criteria with limited algorithmic changes. These findings also showed that SEval-NAS complements training-free NAS methods focused on accuracy, providing broader evaluation across multiple metrics.
Our experiments relied on benchmark-reported hardware metrics, which may differ from measurements on actual devices. This limitation can be mitigated by deploying a lightweight SEval-NAS in an on-device NAS setting. Developing such a system and exploring additional FreeREA thresholds remain future work.

\section*{Acknowledgement}
This work has been supported by NSERC Discovery Grant No RGPIN 2025-00129.

\printbibliography

\appendix
\section{Appendix}\label{appd:A}
\subsection{Ablation Study 1: Evaluation among T5 encoders across Bi-Objective Setups)}\label{appd:A.1}

\begin{figure}[htbp]
    \centering
    \begin{subfigure}[b]{0.4\linewidth}
        \centering
        \includegraphics[width=\linewidth]{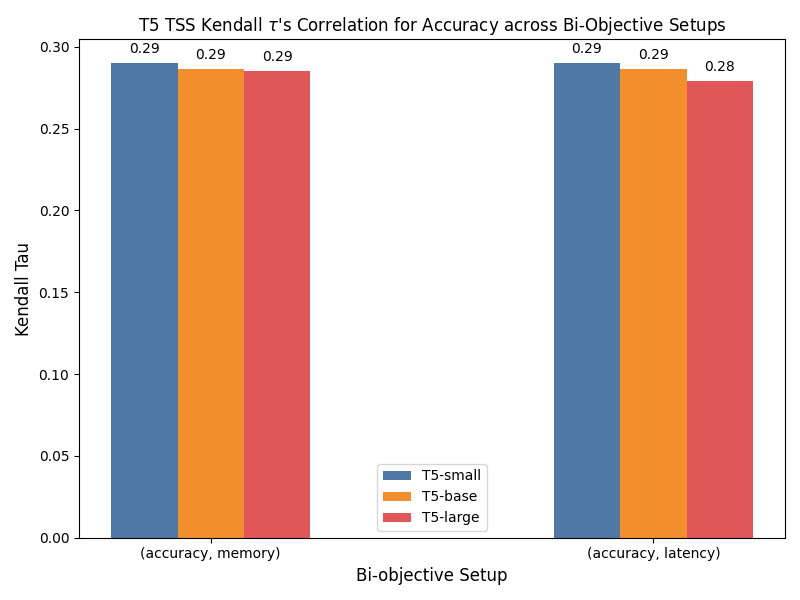}
        \caption{NATS-Bench TSS}
    \end{subfigure}
    \begin{subfigure}[b]{0.4\linewidth}
        \centering
        \includegraphics[width=\linewidth]{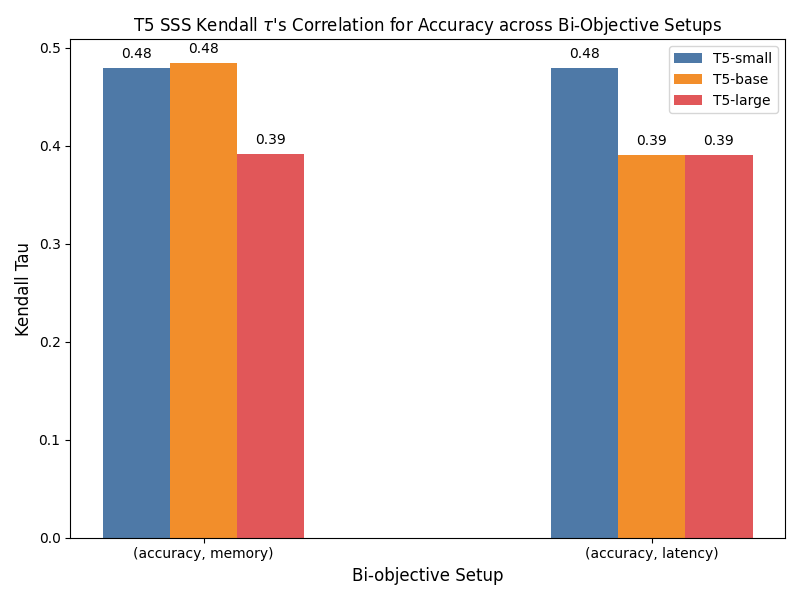}
        \caption{NATS-Bench SSS}
    \end{subfigure}
    \caption{Kendall's $\tau$ correlation for predicted vs true accuracy among encoders of \textcolor{magenta}{T5-small}, \textcolor{magenta}{T5-base}, and \textcolor{magenta}{T5-large} on \textcolor{blue}{NATS-Bench TSS} and \textcolor{green}{NATS-Bench SSS} \textit{(accuracy, latency)} and \textit{(accuracy, memory)} bi-objectives.}
    \label{fig:encoder_kendall}
\end{figure}
\FloatBarrier
\subsection{Ablation Study 2: Feasibility Testing (Evaluation on NATS-Bench)}\label{appd:A.2}

\begin{figure}[htbp]
    \centering
    \begin{minipage}{0.05\linewidth}
        \rotatebox{90}{\fontfamily{phv}\selectfont\large\textbf{\textcolor{violet}{Memory}}} 
    \end{minipage}\hspace{0pt}
    \begin{minipage}{0.92\linewidth}
        \begin{subfigure}[b]{0.31\linewidth}
            \centering
            \includegraphics[width=\linewidth]{figs/plots/nats_bench/tss_ImageNet16-120_memory.png}
            \caption{T5-small}
        \end{subfigure}
        \hfill
        \begin{subfigure}[b]{0.31\linewidth}
            \centering
            \includegraphics[width=\linewidth]{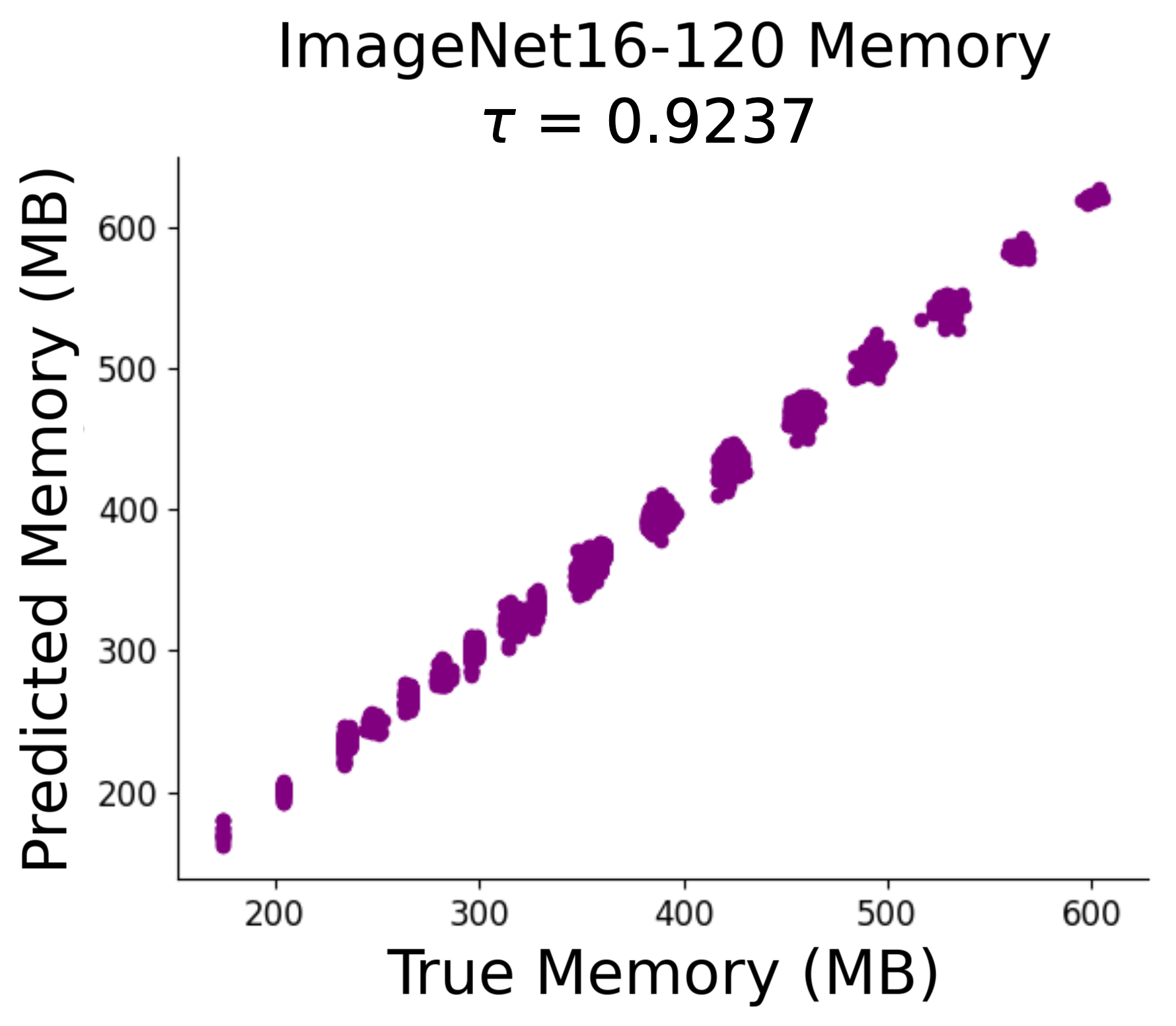}
            \caption{T5-base}
        \end{subfigure}
        \hfill
        \begin{subfigure}[b]{0.31\linewidth}
            \centering
            \includegraphics[width=\linewidth]{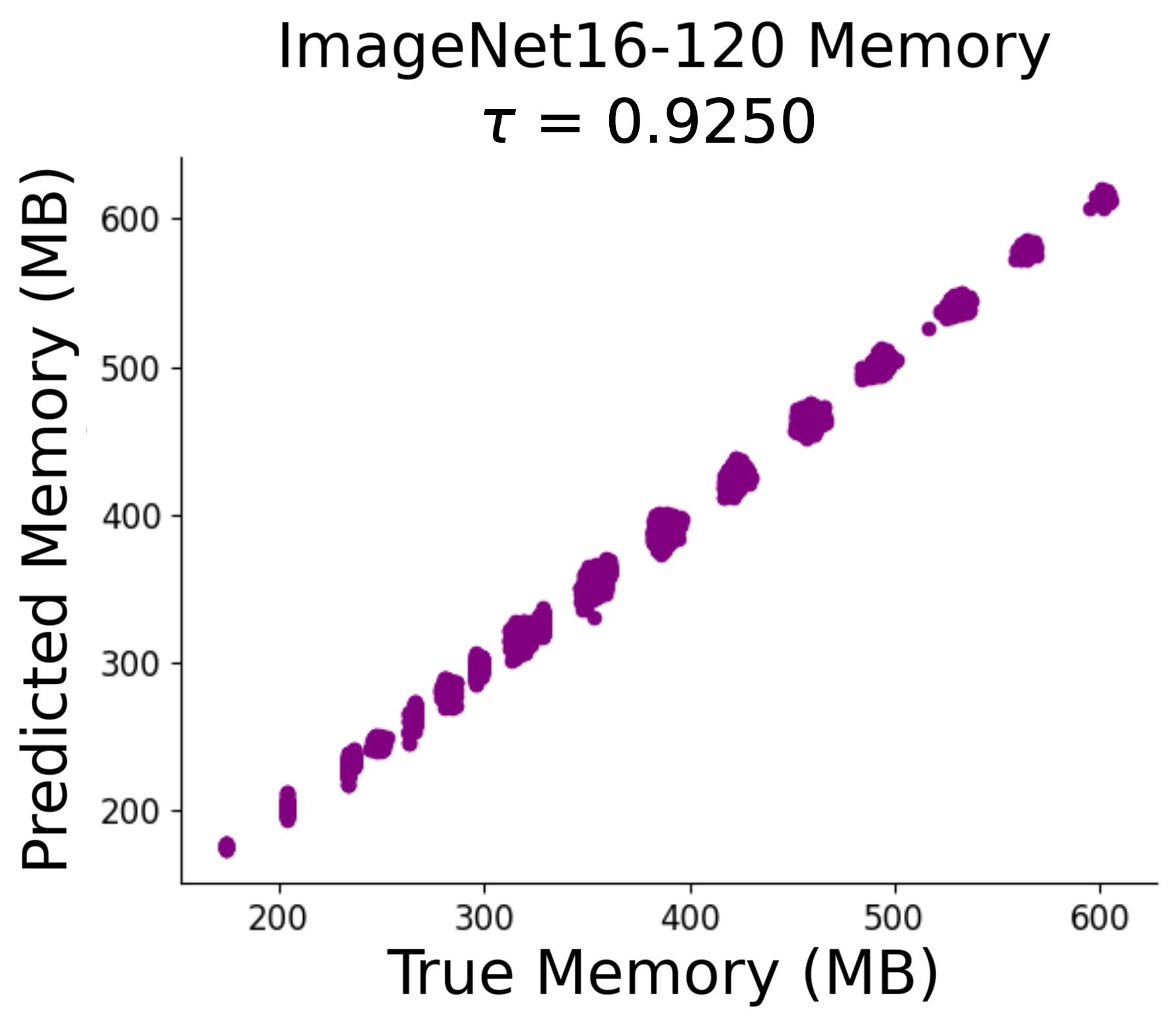}
            \caption{T5-large}
        \end{subfigure}
    \end{minipage}
    \begin{minipage}{0.05\linewidth}
        \rotatebox{90}{\fontfamily{phv}\selectfont\large\textbf{\textcolor{orange}{Latency}}} 
    \end{minipage}\hspace{0pt}
    \begin{minipage}{0.92\linewidth}  
        \begin{subfigure}[b]{0.31\linewidth}
            \centering
            \includegraphics[width=\linewidth]{figs/plots/nats_bench/tss_ImageNet16-120_latency.png}
            \caption{T5-small}
        \end{subfigure}
        \hfill
        \begin{subfigure}[b]{0.31\linewidth}
            \centering
            \includegraphics[width=\linewidth]{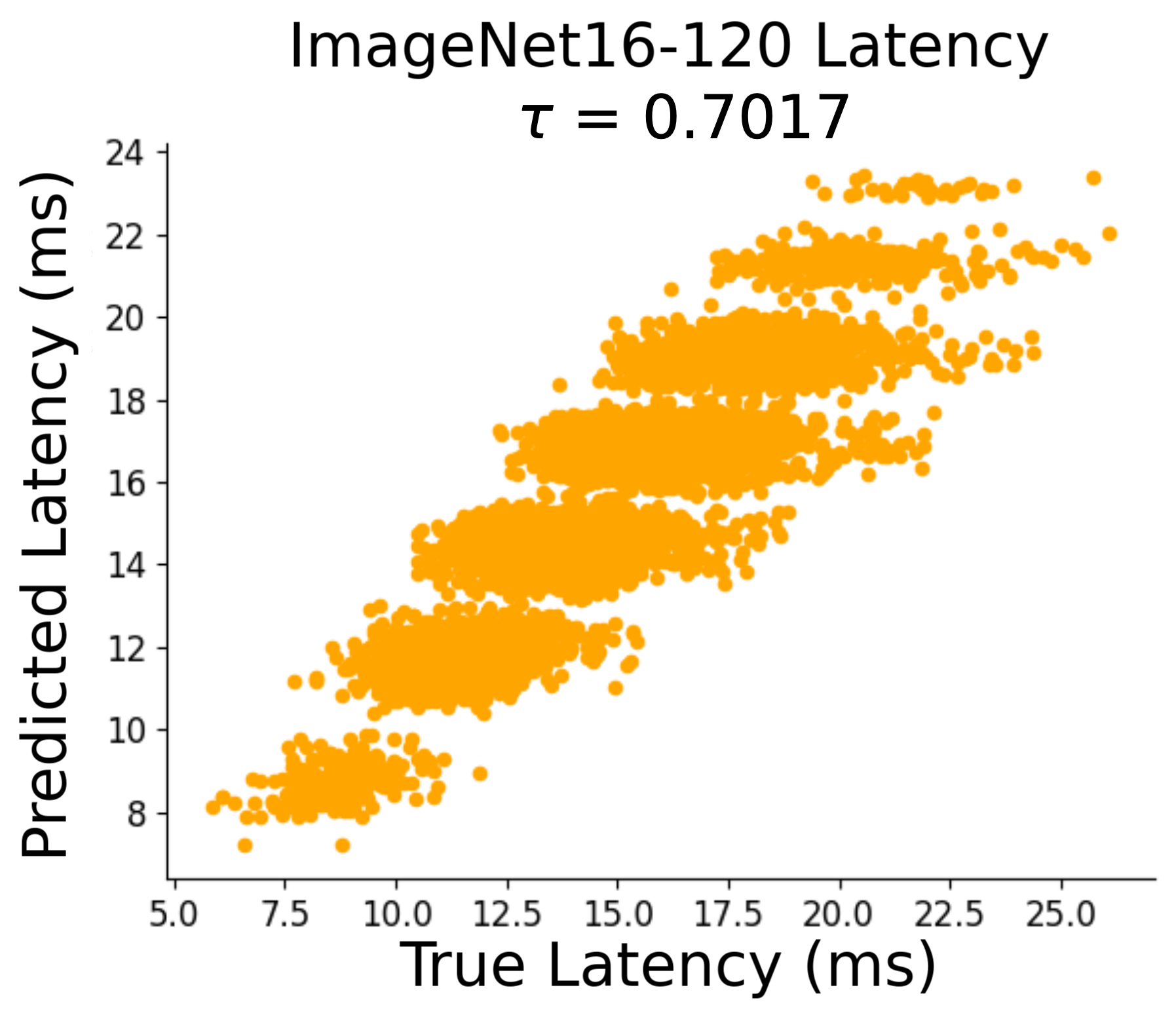}
            \caption{T5-base}
        \end{subfigure}
        \hfill
        \begin{subfigure}[b]{0.31\linewidth}
            \centering
            \includegraphics[width=\linewidth]{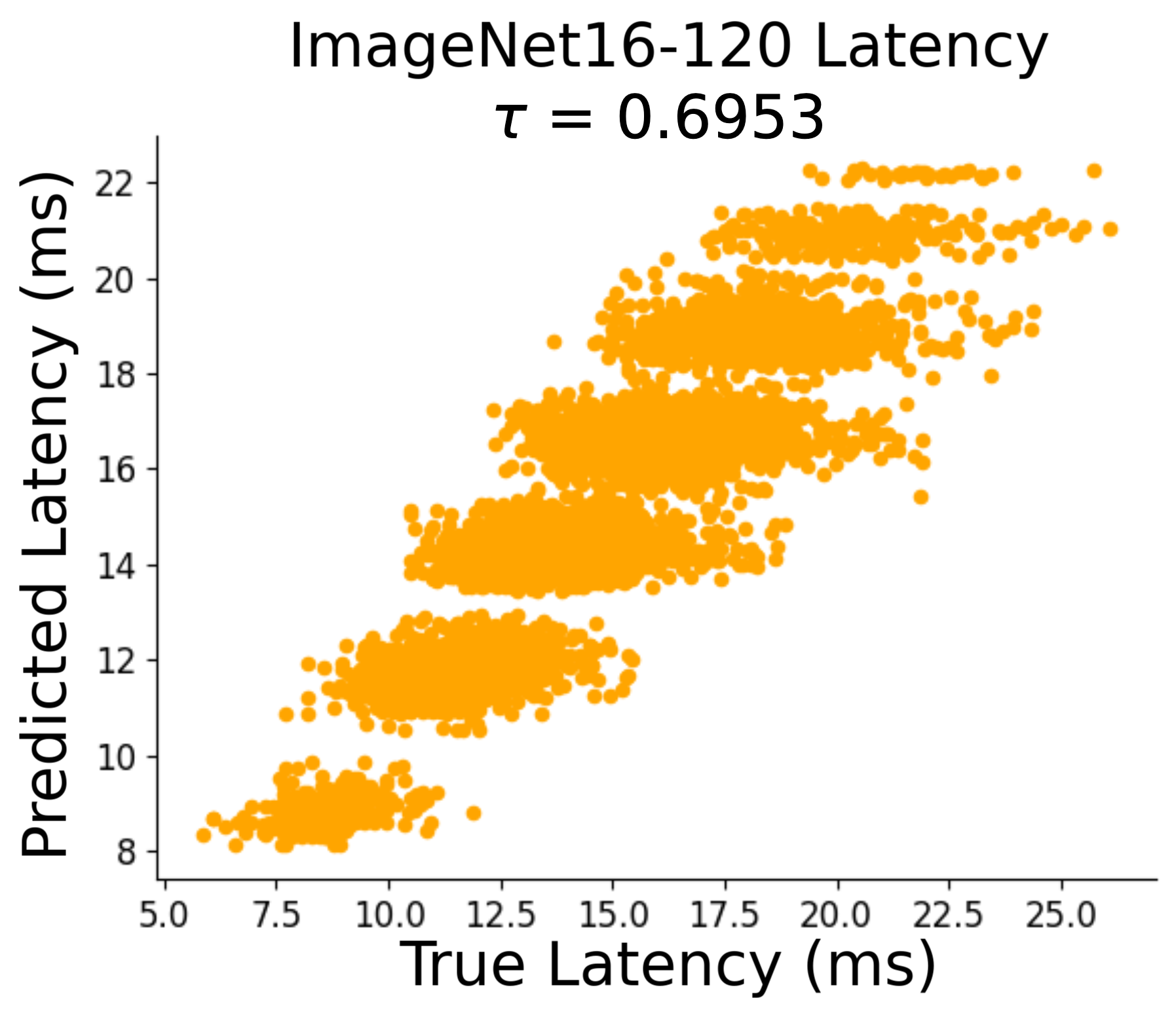}
            \caption{T5-large}
        \end{subfigure}
    \end{minipage}
    \caption{Plots of predicted vs true hardware cost of NATS-Bench TSS architectures for performance metrics reported on \textcolor{magenta}{T5-small}, \textcolor{magenta}{T5-base}, and \textcolor{magenta}{T5-large}. The strength of correlation increases as $\tau$ approaches 1.}
    \label{fig:encoders_tss}
\end{figure}

\begin{figure}[htbp]
    \centering
    \begin{minipage}{0.05\linewidth}
        \rotatebox{90}{\fontfamily{phv}\selectfont\large\textbf{\textcolor{violet}{Memory}}} 
    \end{minipage}\hspace{0pt}
    \begin{minipage}{0.92\linewidth}
        \begin{subfigure}[b]{0.31\linewidth}
            \centering
            \includegraphics[width=\linewidth]{figs/plots/nats_bench/sss_ImageNet16-120_memory.png}
            \caption{T5-small}
        \end{subfigure}
        \hfill
        \begin{subfigure}[b]{0.31\linewidth}
            \centering
            \includegraphics[width=\linewidth]{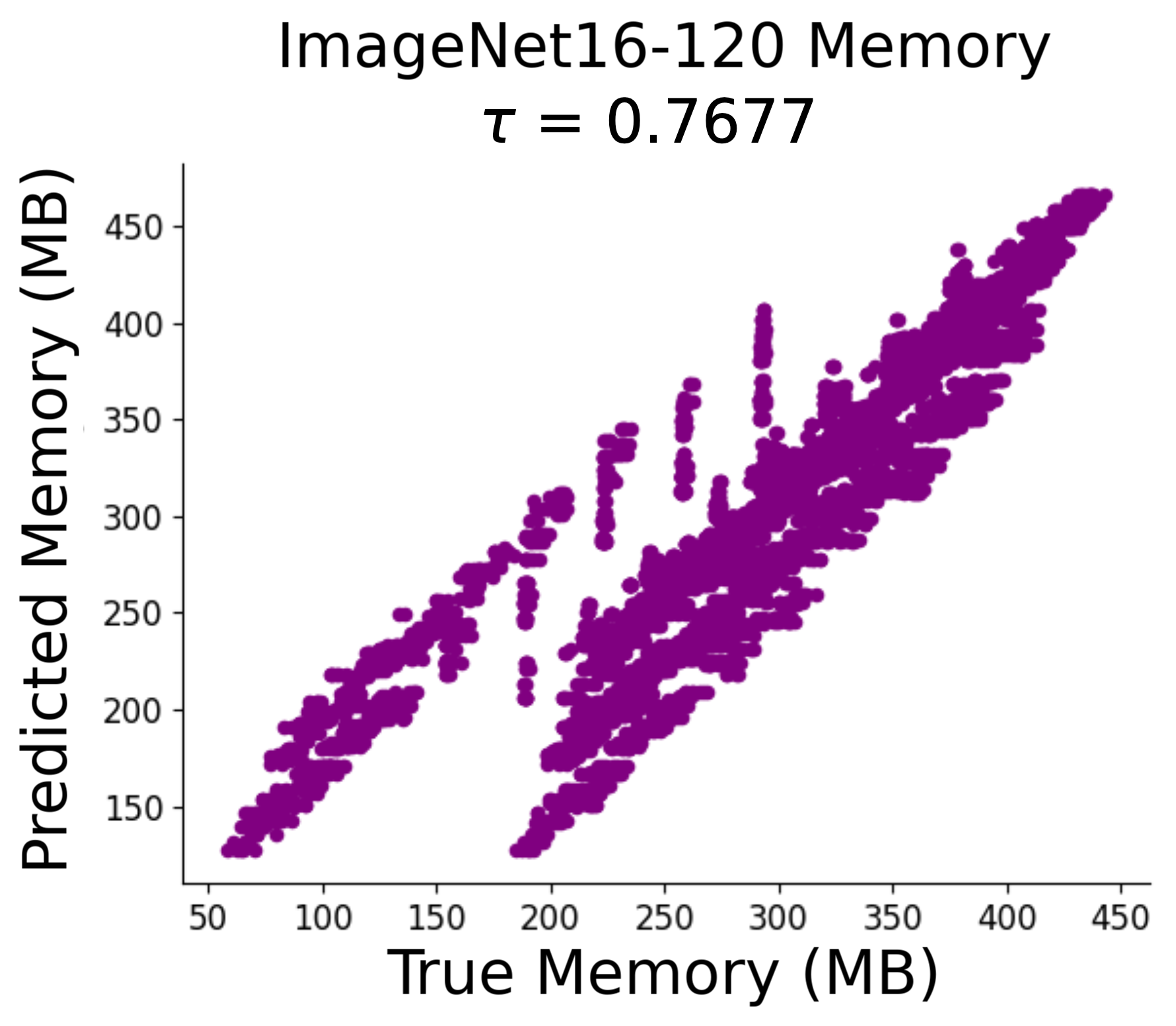}
            \caption{T5-base}
        \end{subfigure}
        \hfill
        \begin{subfigure}[b]{0.31\linewidth}
            \centering
            \includegraphics[width=\linewidth]{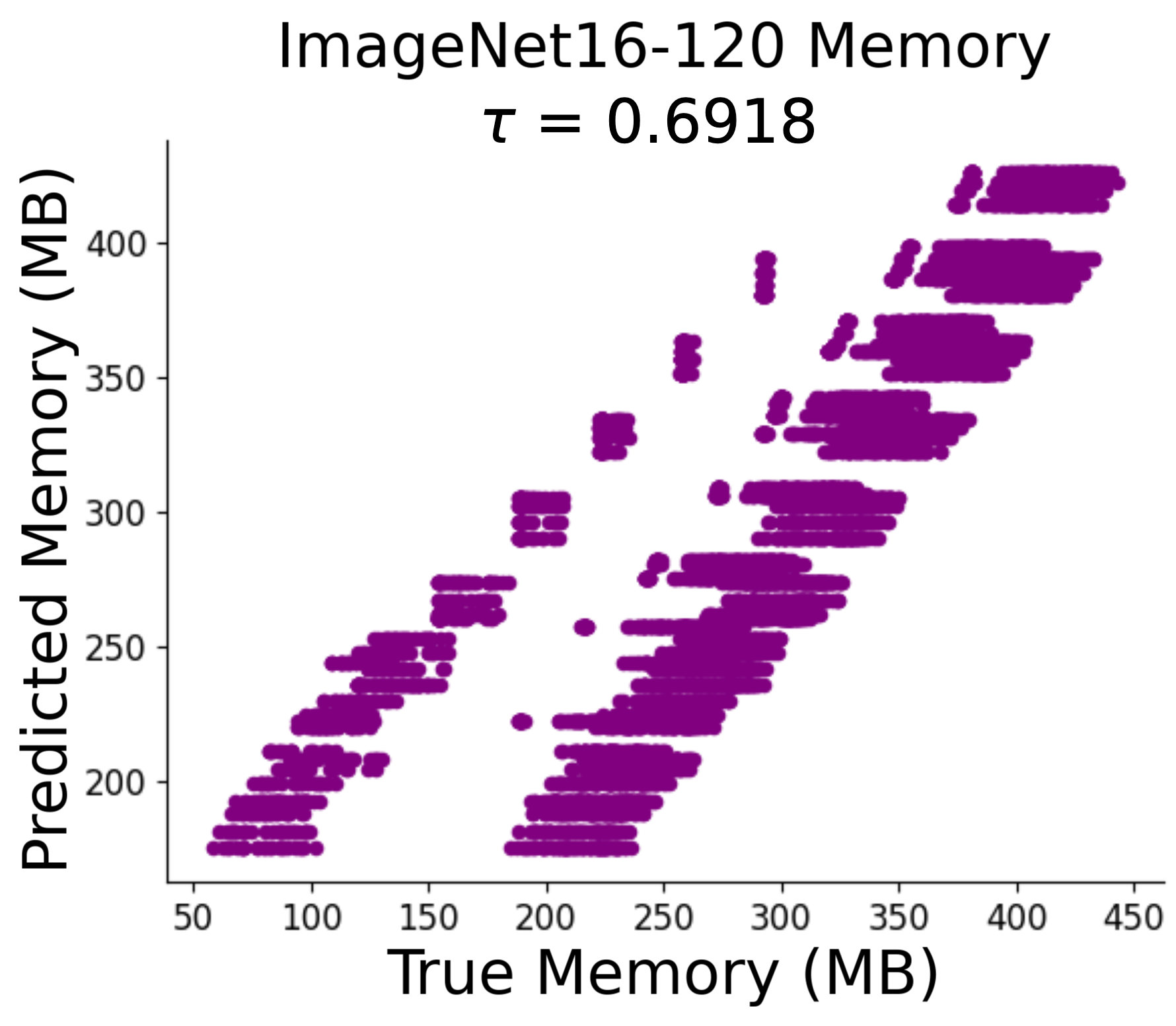}
            \caption{T5-large}
        \end{subfigure}
    \end{minipage}
    \begin{minipage}{0.05\linewidth}
        \rotatebox{90}{\fontfamily{phv}\selectfont\large\textbf{\textcolor{orange}{Latency}}} 
    \end{minipage}\hspace{0pt}
    \begin{minipage}{0.92\linewidth}  
        \begin{subfigure}[b]{0.31\linewidth}
            \centering
            \includegraphics[width=\linewidth]{figs/plots/nats_bench/sss_ImageNet16-120_latency.png}
            \caption{T5-small}
        \end{subfigure}
        \hfill
        \begin{subfigure}[b]{0.31\linewidth}
            \centering
            \includegraphics[width=\linewidth]{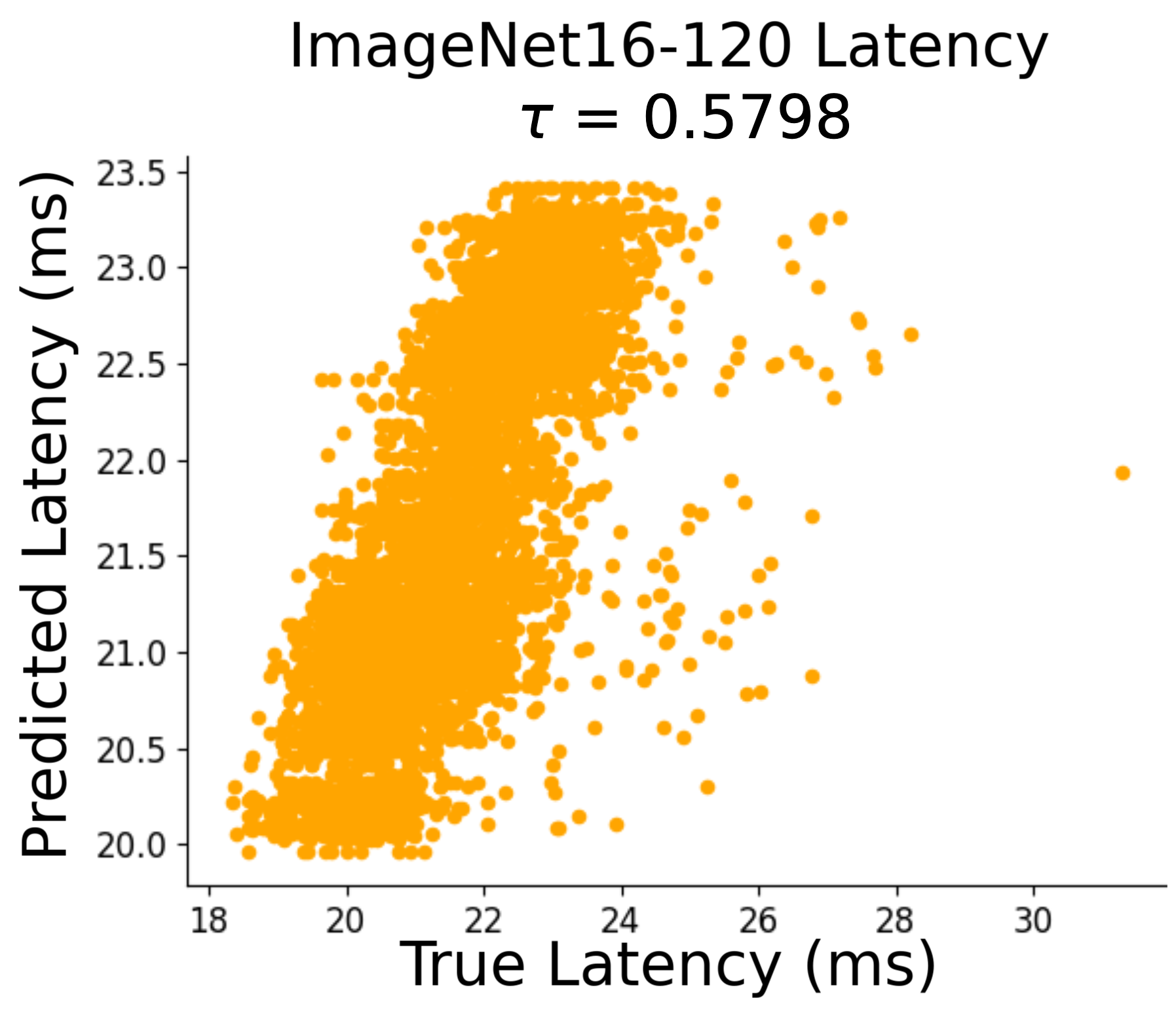}
            \caption{T5-base}
        \end{subfigure}
        \hfill
        \begin{subfigure}[b]{0.31\linewidth}
            \centering
            \includegraphics[width=\linewidth]{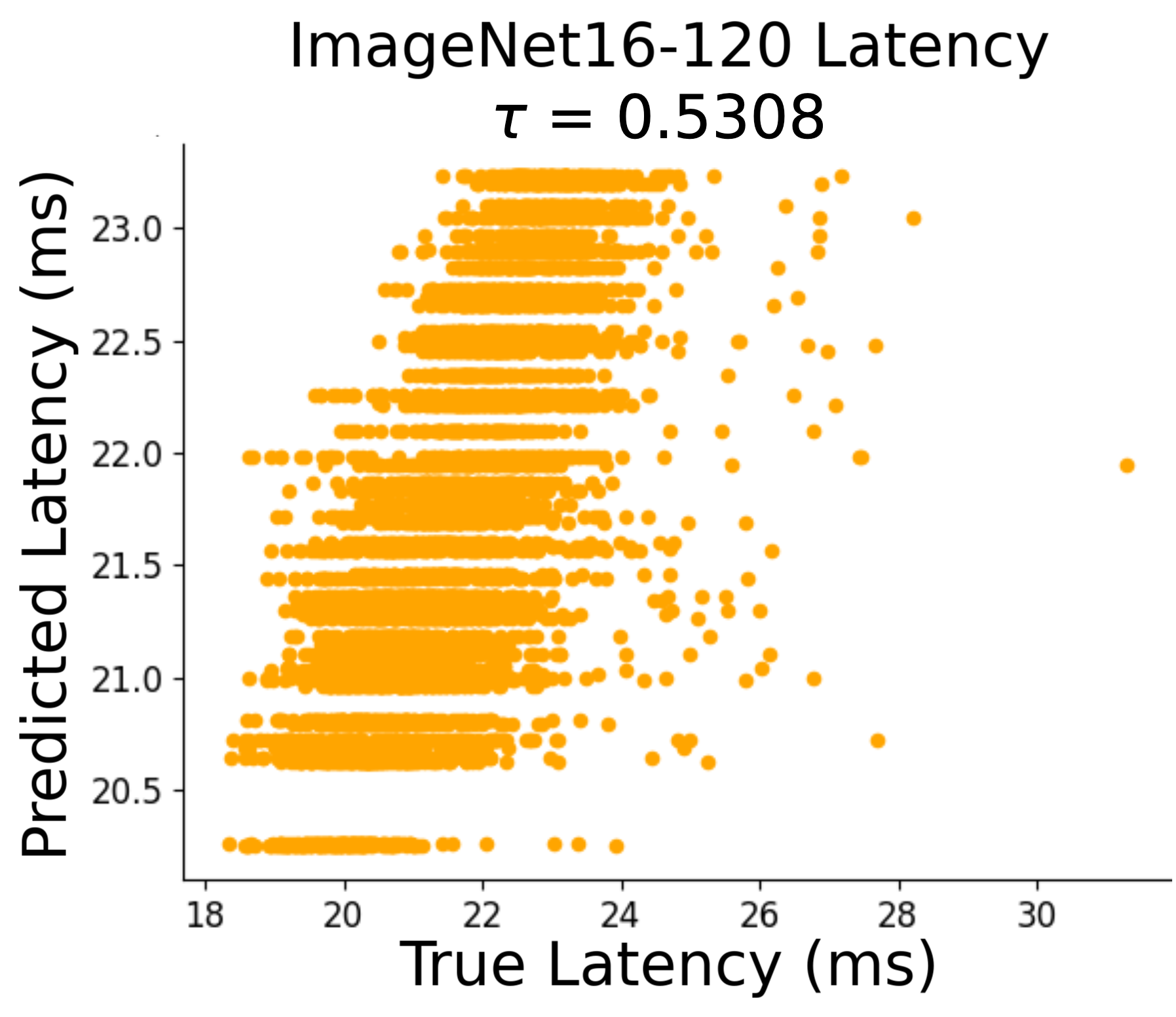}
            \caption{T5-large}
        \end{subfigure}
    \end{minipage}
    \caption{Plots of predicted vs true hardware cost of NATS-Bench SSS architectures for performance metrics reported on \textcolor{magenta}{T5-small}, \textcolor{magenta}{T5-base}, and \textcolor{magenta}{T5-large}. The strength of correlation increases as $\tau$ approaches 1.}
    \label{fig:encoders_sss}
\end{figure}

\newpage
\subsection{Ablation Study 3: Predicting Hardware Cost (Evaluation on HW-NAS-Bench)} \label{appd:A.3}

\begin{figure}[htbp]
    \centering
    \begin{subfigure}[b]{0.32\linewidth}
        \centering
        \includegraphics[width=\linewidth]{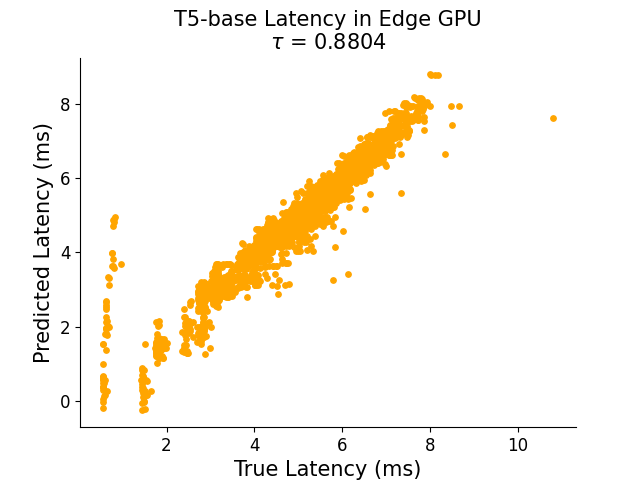}
        \caption{Edge GPU}
    \end{subfigure}
    \hfill
    \begin{subfigure}[b]{0.32\linewidth}
        \centering
        \includegraphics[width=\linewidth]{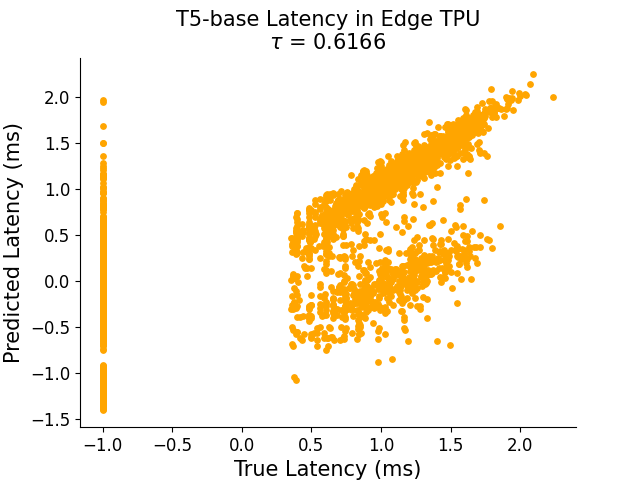}
        \caption{Edge TPU}
    \end{subfigure}
    \hfill
    \begin{subfigure}[b]{0.32\linewidth}
        \centering
        \includegraphics[width=\linewidth]{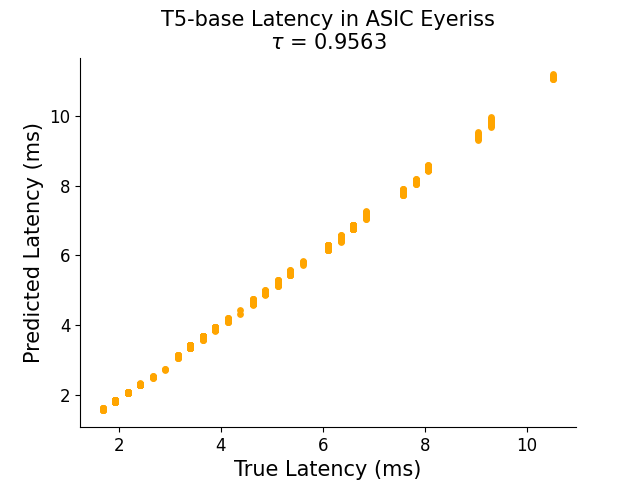}
        \caption{ASIC Eyeriss}
    \end{subfigure}
    
    \begin{subfigure}[b]{0.32\linewidth}
        \centering
        \includegraphics[width=\linewidth]{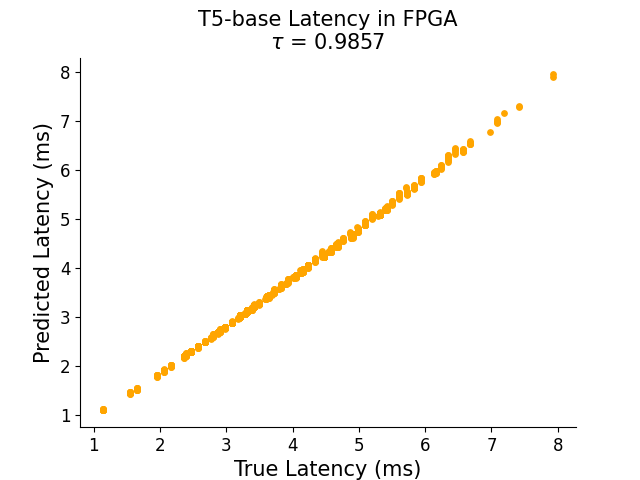}
        \caption{FPGA}
    \end{subfigure}
    \hfill
    \begin{subfigure}[b]{0.32\linewidth}
        \centering
        \includegraphics[width=\linewidth]{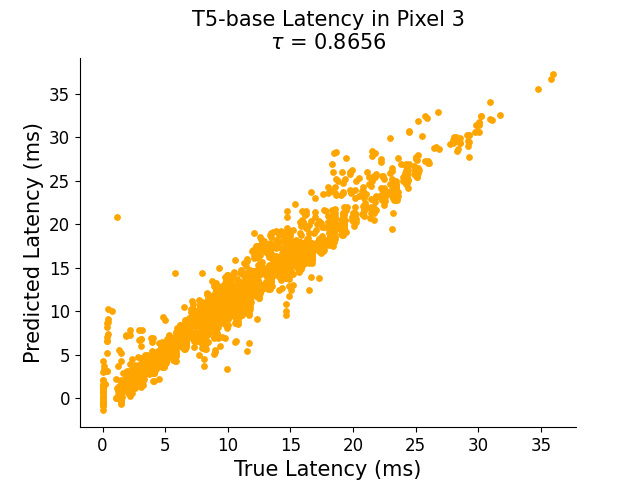}
        \caption{Pixel 3}
    \end{subfigure}
    \hfill
    \begin{subfigure}[b]{0.32\linewidth}
        \centering
        \includegraphics[width=\linewidth]{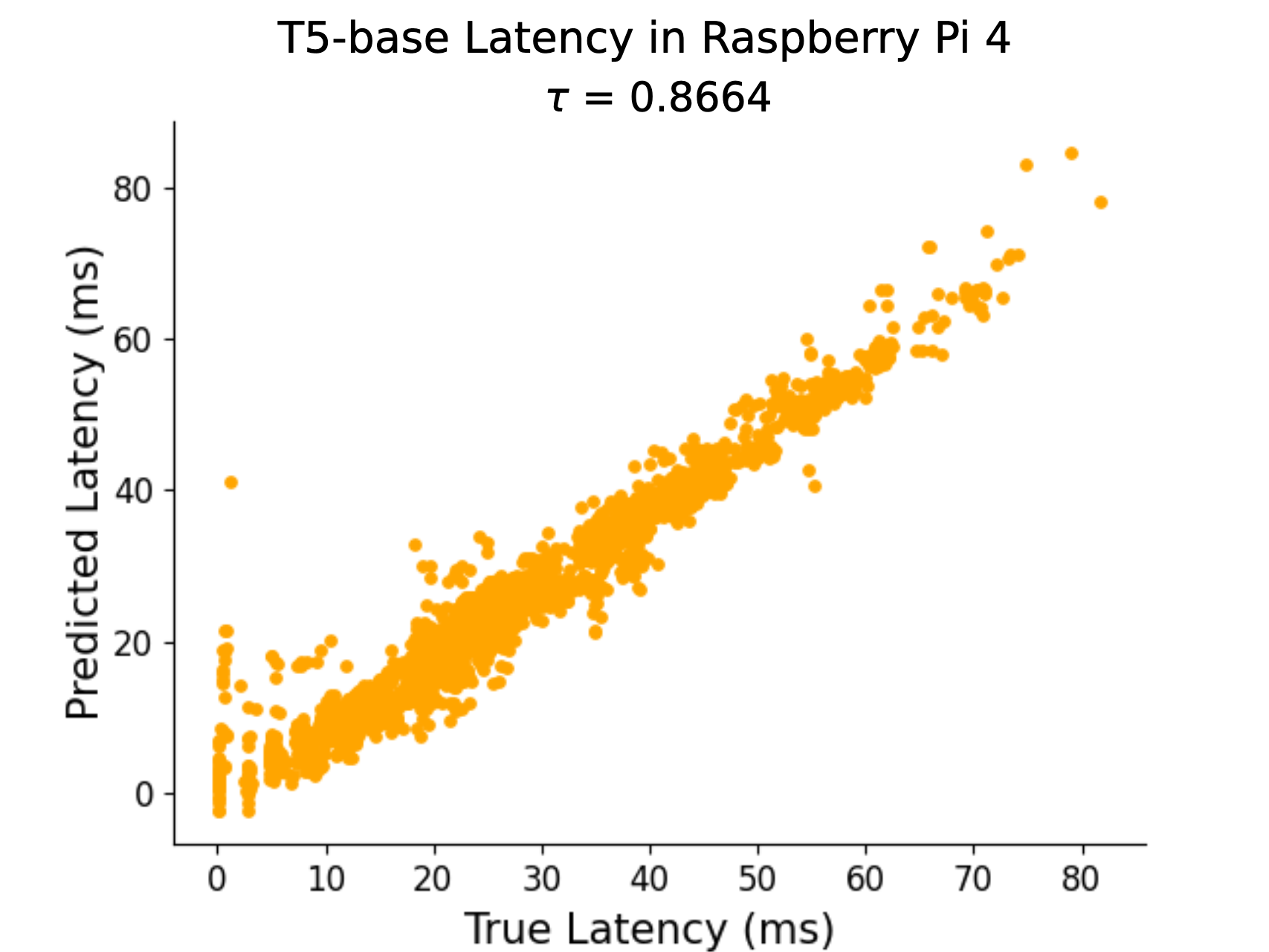}
        \caption{Raspberry Pi 4}
    \end{subfigure}

    \caption{Plots of predicted vs true latency for NAS-Bench-201 architectures with \textcolor{magenta}{T5-base} encoder for 6 edge devices reported in the HW-NAS-Bench benchmark. The strength of correlation increases as $\tau$ approaches 1.}
    \label{fig:plots_hwnas_base}
\end{figure}
\FloatBarrier

\begin{figure}[htbp]
    \centering
    \begin{subfigure}[b]{0.32\linewidth}
        \centering
        \includegraphics[width=\linewidth]{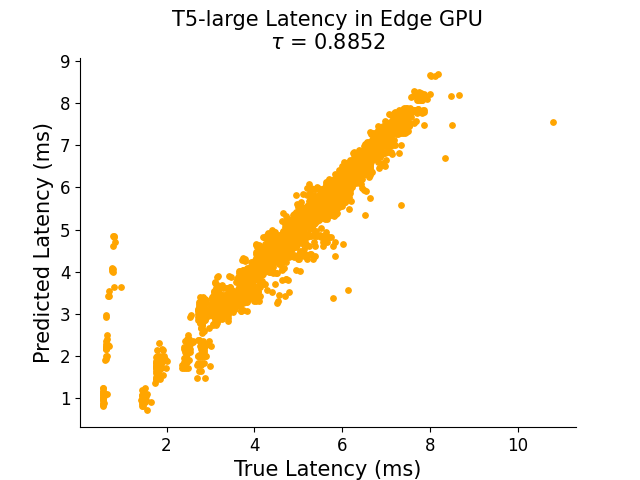}
        \caption{Edge GPU}
    \end{subfigure}
    \hfill
    \begin{subfigure}[b]{0.32\linewidth}
        \centering
        \includegraphics[width=\linewidth]{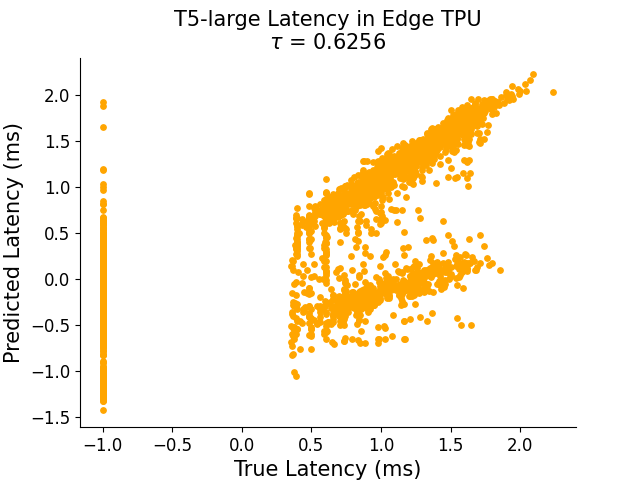}
        \caption{Edge TPU}
    \end{subfigure}
    \hfill
    \begin{subfigure}[b]{0.32\linewidth}
        \centering
        \includegraphics[width=\linewidth]{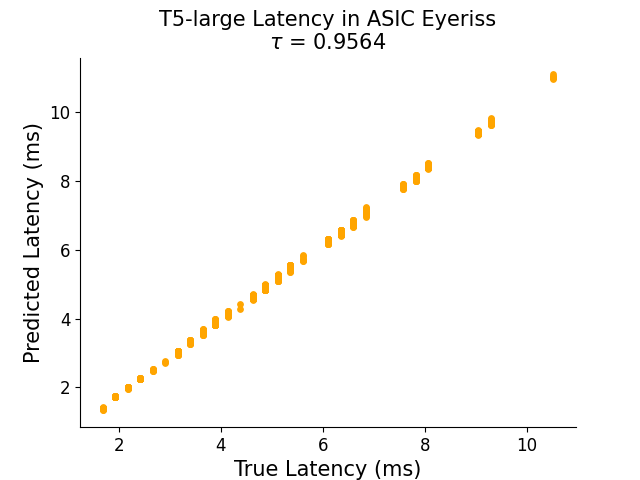}
        \caption{ASIC Eyeriss}
    \end{subfigure}
    
    \begin{subfigure}[b]{0.32\linewidth}
        \centering
        \includegraphics[width=\linewidth]{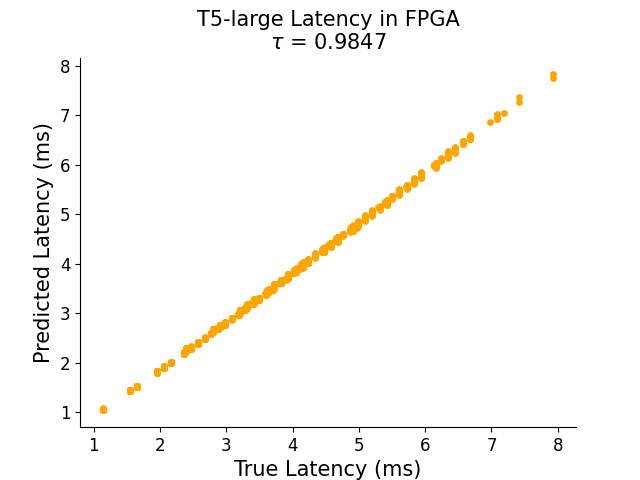}
        \caption{FPGA}
    \end{subfigure}
    \hfill
    \begin{subfigure}[b]{0.32\linewidth}
        \centering
        \includegraphics[width=\linewidth]{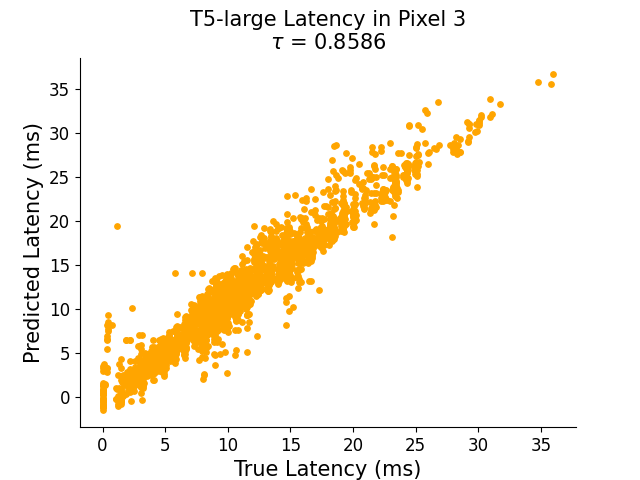}
        \caption{Pixel 3}
    \end{subfigure}
    \hfill
    \begin{subfigure}[b]{0.32\linewidth}
        \centering
        \includegraphics[width=\linewidth]{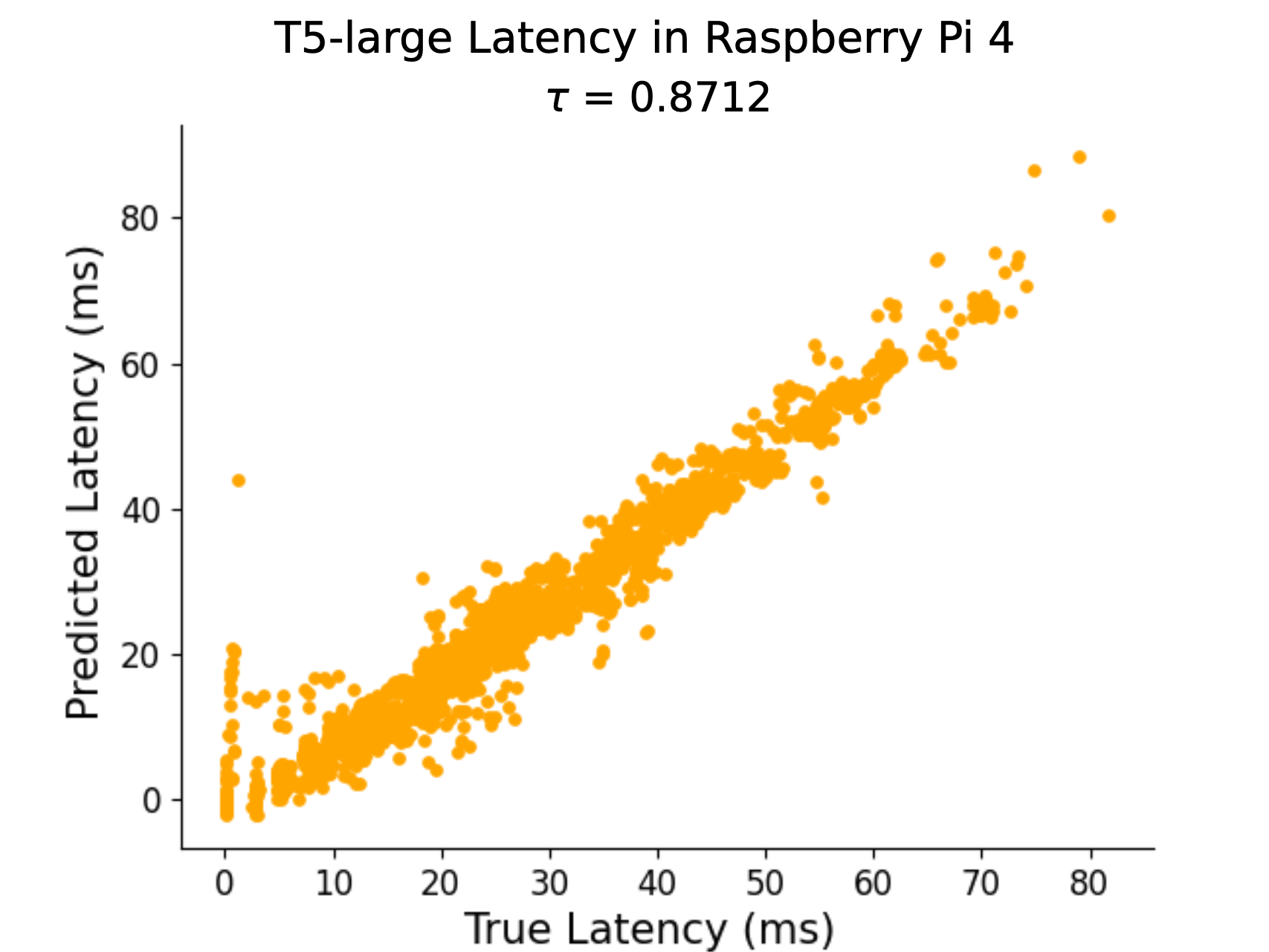}
        \caption{Raspberry Pi 4}
    \end{subfigure}
    \caption{Plots of predicted vs true latency for NAS-Bench-201 architectures with \textcolor{magenta}{T5-large} encoder for 6 edge devices reported in the HW-NAS-Bench benchmark. The strength of correlation increases as $\tau$ approaches 1.}
    \label{fig:plots_hwnas_large}
\end{figure}
\FloatBarrier
\end{document}